%% file: main.tex
\documentclass{tlp}

\usepackage{amsmath}
\usepackage{graphicx}
\usepackage{multirow}

\usepackage{graphicx} % Required for inserting images
\usepackage{amssymb}
\usepackage{hyperref}
\usepackage{fancyvrb}
\usepackage{booktabs}
\usepackage{float} % For custom float environments
\usepackage{caption}
\usepackage{cases}
\usepackage{rotating}
\usepackage{subcaption}
\usepackage{lmodern}
\usepackage{enumitem}

\usepackage{amssymb}
\usepackage{mathtools}
\usepackage{stmaryrd}
\usepackage{booktabs}

\usepackage[ruled,linesnumbered]{algorithm2e} % Enables the writing of pseudo code.
\DontPrintSemicolon

\usepackage{listings}
\lstdefinestyle{mystyle}{
    basicstyle=\ttfamily\footnotesize,
    breakatwhitespace=false,
    breaklines=true,
    captionpos=b,
    keepspaces=true,
    showspaces=false,
    showstringspaces=false,
    showtabs=false,
    tabsize=2,
}
\SetKwInput{KwParam}{Parameter}
\SetKwProg{Fn}{def}{\string:}{}
\newcommand{\qlearning}{\texorpdfstring{$Q$-learning}{Q-learning}}

\newcommand{\reals}{\ensuremath{\mathbb R}}

\newcommand{\natz}{\ensuremath{\mathbb N _ 0}}
\newcommand{\nafsym}{\ensuremath{\textit{not}}}
\newcommand{\naf}{\ensuremath{\nafsym\ }}
\newcommand{\powerset}[1]{\ensuremath{2^{#1}}}

\newcommand{\rv}[1]{\ensuremath{\mathsf{#1}}}

\newcommand{\ex}{\ensuremath{\mathbb E}}

\newcommand{\var}{\ensuremath{\textit{Var}}}
\newcommand{\func}{\ensuremath{\textit{Func}}}
\newcommand{\pred}{\ensuremath{\textit{Pred}}}
\newcommand{\term}{\ensuremath{\textit{Term}}}

\newcommand{\sign}{\ensuremath{\Sigma=(\func,\pred)}}
\newcommand{\hu}{\ensuremath{\mathcal{U}}}
\newcommand{\hb}{\ensuremath{\mathcal{B}}} % F, Ps+Pa
\newcommand{\hi}{\ensuremath{\mathcal{I}}}
\newcommand{\rbody}{\ensuremath{b_1, \ldots, b_n}}

\newcommand{\as}{\ensuremath{\mathcal{AS}}}
\newcommand{\atm}[1]{\ensuremath{\textbf{#1}}}
\newcommand{\trm}[1]{\ensuremath{\textit{#1}}}
\newcommand{\normrule}{\ensuremath{h~\gets~\rbody.}}
\newcommand{\normgoal}{\ensuremath{\gets~\rbody.}}

\newcommand{\defined}{\doteq}
\newcommand{\fdef}{:}
\newcommand{\rmdpp}{\ensuremath{S,A,R,\Psi,p,\gamma}}

\newcommand{\rmdp}{\ensuremath{(\rmdpp)}}
\newcommand{\rmdps}{\ensuremath{(\func, \pred _S \cup \pred _A)}}

\newcommand{\abs}[1]{\ensuremath{\hat{#1}}}
\newcommand{\rva}[1]{\ensuremath{\abs {\rv #1}}}
% ...

\newcommand{\cc}[1]{\ensuremath{ {\llbracket {#1} \rrbracket_{\hat C}} }}
\newcommand{\ccagnd}[2]{\ensuremath{ {\llbracket {#1} \rrbracket _{\hat C} ^{#2}} }}
\newcommand{\ccs}{\ensuremath{ {\abs s} }}
\newcommand{\cca}{\ensuremath{ {\abs a} }}
\newcommand{\ccas}{\ensuremath{ {\abs A} }}
\newcommand{\carcass}{\ensuremath{\langle ( \ccs_1, \ccas_{\ccs_1}), \ldots, ( \ccs_n, \ccas_{\ccs_n}) \rangle }}
\newcommand{\sldnf}{\ensuremath{\vdash_\text{SLDNF}}}
\newcommand{\cov}{\textit{Cov}}

\let\proof\relax

\usepackage{amsthm}
\makeatletter
\newtheoremstyle{tplprunin}%
  {2pt}   % Space above
  {2pt}   % Space below
  {\normalfont} % Body font
  {}      % Indent
  {\bfseries} % Head font
  {.}     % Punctuation
  {0.5em} % Space after head
  {}
\makeatother

\theoremstyle{tplprunin}
\newtheorem{definition}{Definition}
\newtheorem{example}{Example}

\newtheorem{corollary}{Corollary}

\theoremstyle{plain}
\newtheorem{proposition}{Proposition}

\usepackage{newtxtext,newtxmath}

\usepackage{microtype}

\begin{document}

\lefttitle{Bankosegger, Eiter, Oetsch}

\jnlPage{1}{16}
\jnlDoiYr{2026}
\doival{10.1017/xxxxx}

\title[ASP-based Abstractions for Reinforcement Learning]{Answer-Set-Programming-based Abstractions for Reinforcement Learning} %\thanks{funds}}

\begin{authgrp}
\author{ \sn{Rafael} \gn{Bankosegger}$^{1,2}$, \sn{Thomas} \gn{Eiter}$^2$, \gn{Johannes} \sn{Oetsch}$^3$}
\affiliation{%
$^1$ Siemens AG Österreich, Vienna, Austria \\
$^2$ TU Wien (Vienna University of Technology), Austria \\
$^3$ Jönköping University, Sweden \\
% {\{thomas.eiter,tobias.geibinger,nysret.musliu\}@tuwien.ac.at},
\email{rafael-philipp.bankosegger@siemens.com, thomas.eiter@tuwien.ac.at, johannes.oetsch@ju.se} }
\author{
}      
\end{authgrp}

\history{\sub{xx xx xxxx;} \rev{xx xx xxxx;} \acc{xx xx xxxx}}

\maketitle

\begin{abstract}
Reinforcement Learning (RL) enables autonomous agents to learn policies from experience, but realistic problems often involve enormous state spaces, making learning and generalisation challenging. Abstraction and approximation are therefore essential. Relational Reinforcement Learning (RRL) offers a way to reason about objects and their relations, and the CARCASS framework by Martijn van Otterlo demonstrates how logical representations can model Markov Decision Processes (MDPs) in first-order domains. Originally implemented in Prolog, CARCASS leverages domain knowledge to create powerful abstractions. We explore Answer-Set Programming (ASP), which is a rich and, contrary to Prolog, fully declarative modelling language, to realise CARCASS abstractions. We evaluate our ASP-based implementation in case studies of two domains, viz.\ Blocks World and Minigrid. Our results indicate that CARCASS with ASP provides a promising approach to constructing abstractions for RL, especially when domain knowledge is available.%
\setcounter{footnote}{0}\footnote{Our implementation is available at \url{https://github.com/rbankosegger/RLASP-core}. Further material (data, encodings, extended documentation) can be found here: \url{https://www.bankosegger.at/iclp26/}.}
\end{abstract}


%%%%%%%%%%%%%%%%%%%%%%%%%%%%%%%%%%%%%%%%%%%%%%%%%%%%%%%%%%%%%%
\section{Introduction}\label{sec:intro}
%%%%%%%%%%%%%%%%%%%%%%%%%%%%%%%%%%%%%%%%%%%%%%%%%%%%%%%%%%%%%%

% Reinforcement Learning (RL) enables autonomous agents to learn policies from experience but realistic problems often involve enormous state spaces, making learning and generalisation challenging. Abstraction and approximation are therefore essential.
\emph{Reinforcement Learning} (RL)~\citep{sutton_barto_2018} has emerged as a central paradigm in AI, providing a principled framework for agents to learn behaviour through interaction with an environment. By formalising sequential decision-making as a Markov Decision Process (MDP), %~\citep{Puterman94}, 
RL enables agents to improve their behaviour by trial and error, guided by the maximisation of cumulative reward. %~\citep{sutton_barto_2018}. 
Although this paradigm has produced impressive results {with significant practical applications, a challenge that remains is that} realistic problem settings often involve enormous state and action spaces, which render naive learning approaches intractable. Effective techniques for coping with this curse of dimensionality, such as abstraction and approximation, are therefore indispensable. 

Approximations, e.g., Deep Q-Learning~\citep{Mnih15}, aim at learning compact representations of expected long-term rewards, typically using deep neural networks, thereby enabling generalisation across large or continuous state spaces. In practice, however, this often requires large amounts of training data, and learning can be unstable~\cite[Chapter~11.3]{sutton_barto_2018}.

\emph{Relational Reinforcement Learning} (RRL) addresses some of these limitations by using first-order logic to introduce powerful abstractions for reasoning about complex problem domains. The key idea is to extend representations to a first-order setting, allowing environments to be described naturally in terms of objects and their relations~\citep{otterlo_2005}. A major advantage of this approach is that it supports generalisation across similar states and can even transfer learned knowledge to related tasks.

%and the CARCASS framework by Martijn van Otterlo demonstrates how logical representations can model Markov Decision Processes (MDPs) in first-order domains. Originally implemented in Prolog, CARCASS leverages domain knowledge to create powerful abstractions.
One particular approach to RRL, 
%from the logic-programming world 
rooted in logic programming, is the CARCASS framework (short for \emph{Compact Abstraction using Relational Conjunctions for Aggregation of State-action Spaces}) introduced by 
Martijn van Otterlo~(\citeyear{otterlo_2004a}). 
CARCASS aims at abstracting the state-action space by representing abstract states as conjunctions of first-order literals, possibly enriched with background knowledge, and associates them with sets of admissible actions. An abstraction is thus an ordered list of rules of the form \emph{State} $\rightarrow$ \{Action$_1$, \ldots, Action$_n$\}, each capturing a situation described by the state and the  available actions. For example, the rule
\[
\atm{clear}(X), \, \atm{clear}(Y), \, \atm{on}(Y,0) \;\;\rightarrow\;\; \{\atm{move}(X,Y), \, \atm{move}(Y,X)\}
\]
represents an abstract blocks world state in which two blocks $X$ and $Y$ are clear and can be stacked on one another. 
A \emph{policy} (what action to choose when in a given state) can then be learned via, e.g.,
 $Q$-learning~\citep{Watkins1992} over the abstract state-action space.
One of CARCASS’s key advantages is its ability to leverage domain knowledge in the abstraction rules, thereby reducing the effective size of the state–action space and supporting generalisation across related situations. The framework was %originally
realised in Prolog, making use of SLDNF-resolution to handle first-order reasoning over states and actions~\citep{otterlo_2008_phd}.  %otterlo_2004a}. %,otterlo_2008_phd}.

% We explore Answer-Set Programming (ASP) as an expressive and fully declarative modelling language for CARCASS.
%Building on this, we explore \emph{Answer-Set Programming} (ASP)~\citep{brewka_eiter_truszczynski_2011} %%,gebser_2012,lifschitz_2019} 
%as an alternative approach to revisiting CARCASS. ASP is a knowledge representation and reasoning framework with roots in logic programming, featuring a concise yet highly expressive modelling language which, unlike Prolog, is fully declarative in that the ordering of rules and literals in rule bodies does not affect the evaluation result. It supports nonmonotonic inference, reasoning under incomplete knowledge, expressing preferences and optimisation, as well as easy modelling of actions and change. These characteristics, combined with the ability to declaratively encode domain knowledge, make ASP particularly attractive for describing state abstractions in RRL. 

Building on this, we explore Answer-Set Programming (ASP)~\citep{brewka_eiter_truszczynski_2011} as a
rich and fully declarative modelling language for revisiting CARCASS.\footnote{Pure Prolog is declarative in principle, but in practice many implementations rely on procedural features such as negation as failure, cuts, or the ordering of rules to achieve efficiency.} 
ASP is a logic-based knowledge representation and reasoning framework, featuring a concise yet highly expressive modelling language. Unlike Prolog, the meaning of an ASP program is independent of the ordering of rules and of literals in rule bodies. It supports nonmonotonic inference, reasoning under incomplete knowledge, expressing preferences and optimisation, as well as modelling actions and change, making it particularly suitable for describing state abstractions in RRL.  

Reimplementing CARCASS in ASP highlights several practical advantages. Admissible states, transitions, and optimisation criteria can be specified declaratively at a high level, without explicit procedural search strategies. Defaults and integrity constraints are supported naturally, allowing concise and robust encodings that would otherwise require more algorithmic, control-intensive implementations in Prolog. This also enables direct representation of complex reasoning patterns, including partial observability, non-determinism, and preference-based selection, with domain constraints expressed directly in the rules.

%To the best of our knowledge, only two works so far use ASP in this context: a hierarchical framework in which ASP models high-level policies~\citep{mitchener_2022}, and a robot architecture where ASP is applied to identify and eliminate irrelevant atoms in state descriptions, a form of state abstraction~\citep{sridharan-meadows-2018}. However, no ASP-based representation of a full state–action pair abstraction has yet been proposed.

% contribution
In this work, we introduce a general method for encoding abstractions in ASP, which we evaluate using two case studies. The first considers Blocks World, a classical planning problem that involves stacking blocks to reach a desired configuration, and the second MiniGrid, a suite of grid world navigation tasks involving different subtasks such as key collection and door opening to reach a goal. Both domains have state spaces that are infeasibly large without abstraction. 

Our main contribution can thus be summarised as follows:
\smallskip

\begin{enumerate}%[(1)]
\itemsep=2pt
    \item %[(i)] 
    we introduce a general method for encoding CARCASS abstractions in ASP, 
    %demonstrating 
    showing how a fully declarative and expressive modelling language can realise relational state–action abstractions;
    \item %[(ii)] 
    furthermore, we show how to use ASP-based CARCASS abstractions for online learning; and
    \item %[(iii)] 
    we evaluate our ASP-based implementation in two case studies: Blocks World and MiniGrid. The experiments show that, using $Q$-learning on the abstract representations, high-quality policies can be learned consistently. These policies can also be obtained with significantly fewer samples than  
%would be required  
without the abstraction.
\end{enumerate}
\smallskip

In conclusion, our results indicate that CARCASS with ASP provides a promising approach to constructing abstractions for RL, in particular when domain knowledge is available. 

% Roadmap
The rest of this paper is organised as follows. 
%We start with 
After preliminaries in Section~\ref{sec:prel}, we present in Section~\ref{sec:asp-carcass} %presents 
our ASP-based approach for encoding CARCASS abstractions. Sections~\ref{sec:blocks} and~\ref{sec:minigrid} are devoted to %describe the two case studies, 
%Blocks World and MiniGrid, 
the case studies, 
followed by 
%including 
an empirical evaluation with setup, results, and discussion in Section~\ref{sec:eval}, 
which also
%. We touch on the 
touches modelling differences between ASP and Prolog.
%in (Section~\ref{sec:asp_prolog}).
In Section~\ref{sec:rel}, we review related work, and in Section~\ref{sec:concl}, we conclude with 
%a discussion of implications and 
directions for future work.

\section{Preliminaries}\label{sec:prel}
%%%%%%%%%%%%%%%%%%%%%%%%%%%%%%%

We next review relevant concepts from logic programming (SLDNF-resolution and ASP),
reinforcement learning (relational MDP and {\qlearning}), and of the CARCASS framework.
%	Reinforcement learning.
%	Random variables $\rv S_t$, $\rv A_t$, etc.

% ASP (compact and focused on what is needed)
% RL (MDP, Policy, Q-learning concepts (updates, q-table, state space))
% most important: CARCASS by Otawa  

\subsection{Logic Programming}
We assume a first-order language
with constants $c$, functional terms $f (\bar t)$, and predicate atoms
$\atm{p}(\bar t)$  over a PL1-signature 
$\sign$ and a set $\var = \{ V_1, \ldots, V_m \}$  of variables,
where $\bar t = t_1, \ldots, t_n$ is a list of terms matching the arity of $f$, resp., $p$.

%\paragraph{Normal logic programs, SLDNF-resolution.}
{
    A \emph{naf-literal} is an atom $\atm{p}(\bar t)$ or an expression $\naf \atm{p}(\bar t)$ with negation as failure.
}
A \emph{normal rule} is of the form $\normrule$,
where $h$ is an atom and $b_1, \ldots, b_n$ are naf-literals.
A \emph{normal program} $P= \{r_1, \ldots, r_n \}$ is a set of normal rules.
The application of a \emph{substitution} $\theta \fdef \var \to \term$ from a set of variables 
to a set of terms is defined as usual.
Syntactic objects without variables are called \emph{ground}.
Given a program $P$ and a \emph{normal goal} $\normgoal$,
%where $b_1, \ldots, b_n$ are naf-literals,
%the existence of 
an \emph{SLDNF-refutation} of $P \cup \{ \normgoal \}$  
is denoted by $P \sldnf \rbody$ \citep[Chapter~8]{nienhuys_wolf_1997}.

%\paragraph{Herbrand interpretations.}
Based on a signature $\sign$ we denote 
by $\hu(\Sigma)$ the Herbrand universe over $\func$,
by $\hb(\Sigma)$ the Herbrand base over $\hu(\Sigma)$ and $\pred$,
and by $\hi(\Sigma) = \powerset{\hb(\Sigma)} $ the set of Herbrand interpretations.
%An interpretation $I \in \hi(\Sigma)$ is to be read such that 
In an interpretation $I \in \hi(\Sigma)$, an atom $\atm{p}(\bar t)$ is \emph{true} if $\atm{p}(\bar t) \in I$ and
\emph{false} if $\atm{p}(\bar t) \not \in I$.

%\paragraph{Answer-Set Programming (ASP).}
We consider ASP programs as defined in the \emph{ASP-Core-2 input language format}~\citep{calimeri_gebser_ianni_schaub_2020},
with the usual features such as strong negation, disjunctive rule heads, and optimisation.
In particular, our encodings make use of choice rules, aggregates, and weak constraints.
The set of all answer sets of a program $P$ is denoted by 
%$\as(P) = \{ I, J, \ldots \}$.
$\as(P)$.

\subsection{Reinforcement Learning}
Reinforcement Learning \citep{sutton_barto_2018}
is a discrete-time, stochastic control process 
% control process?
governed by the dynamics of a \emph{task environment}
and by the actions of a learning \emph{agent}.
At every time point $t$, the agent perceives the current \emph{state} $\rv S_t$ of the environment
and takes an \emph{action} $\rv A_t$.
%This choice is made based on the agents current policy $\rv \Pi_t$.
%After advancing the clock,
The environment transitions to a new state $\rv S_{t+1}$ based on the effects of the action
and a \emph{reward} $\rv R_{t+1}$ is obtained.
The result is a history of interactions
%$\rv S_0$, $\rv A_0$, $\rv R_1$, $\rv S_1$, $\rv A_1$, $\rv R_2$, $\rv S_2$, $\ldots$
$\rv H = (\rv S_0, \rv A_0, \rv R_1, \rv S_1, \rv A_1, \rv R_2, \rv S_2, \ldots )$.
%TODO:
The performance of the agent is measured in terms of the \emph{$\gamma$-discounted return} $\rv G_t \defined \sum_{k=0}^{\infty} \gamma^{k} \rv R_{t+k+1}$.

%\paragraph{Relational environments.}
The task environment is formalised as a \emph{relational Markov decision process (RMDP)}~{\citep[p.~168]{otterlo_2008_phd}}.
Given a PL1-signature $\Sigma = \rmdps$, an RMDP consists of a set $S \subseteq \hi ((\func, \pred_S))$ of states, 
%where each state is a Herbrand interpretation,
which are Herbrand interpretations, a set $A \subseteq \hb ((\func, \pred_A))$ of actions, where each action is an atom, and a set  $R \subseteq \reals$ of rewards. For every state $s \in S$, a set $A_s$ of admissible actions is defined. The set of admissible state-action pairs is $\Psi \defined \{ (s,a) \mid s \in S, a \in A_s \}$. % maybe cite ravindran_2004_phd 
The \emph{dynamics} of the environment are defined by the probability $p(s', r \mid s, a)$
of transitioning to state $s'$ with a reward of $r$ after performing  action $a$ in state $s$,
i.e. $\rv S_{t+1}, \rv R_{t+1} \sim p(\ \cdot \mid \rv S_t, \rv A_t)$.
With the discount rate $\gamma$, an RMDP is thus a tuple $M=\rmdp$.
We further define the logic programs
	$P_s \defined \{\ \atm{p}(\bar t). \mid \atm{p}(\bar t) \in s  \ \}$
	and 
	$P_{A_s} \defined \{\ \atm{p}(\bar t). \mid \atm{p}(\bar t) \in A_s \ \}$.
%			%is a probability distribution that
%			%	with 
%			%	\[\sum_{s' \in S} \sum_{r \in R} p(s',r \mid s,a) = 1 
%			%	\quad \forall (s,a) \in \Psi
%			%	\]
%			defines the \emph{dynamics} of the environment, i.e.,
%			the probability of transitioning to state $s'$ with a reward of $r$ after performing  action $a$ in state $s$;
%			and 

\begin{figure}[t]
	\begin{center}
	\begin{minipage}{0.3\textwidth}
	\includegraphics[height=1.7cm]{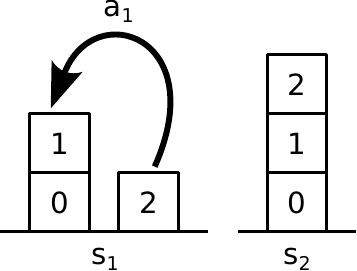}
	\end{minipage}
	%\hspace{1em}
	\begin{minipage}{0.4\textwidth}
	\scriptsize
	\begin{align*}
		s_1 & = \left\{ \begin{array}{l}
			\atm{on}(0, table), \atm{on}(1, 0), \atm{on}(2, table), \\
			\atm{goal}(0,table), \atm{goal}(1,0), \atm{goal}(2,1)
			\end{array}\right\}\\
		a_1 &= \atm{move} (2,1) \\
		s_2 & = \left\{ \begin{array}{l}
			\atm{on}(0, table), \atm{on}(1, 0), \atm{on}(2, 1), \\
			\atm{goal}(0,table), \atm{goal}(1,0), \atm{goal}(2,1)
			\end{array}\right\}
	\end{align*}
	\end{minipage}
	\end{center}
	\caption{A $3$-blocks world RMDP.
            Taking action $a_1$ in state $s_1$ causes a transition to $s_2$.
            %In state $s_1$, the agent takes action $a_1$ and the resulting next state is $s_2$.
	}
	\label{fig:ex-rmdp-bw}
	\end{figure}

\begin{example}[Blocks World]\label{ex:rmdp-bw}
Consider a $3$-blocks world setting \citep{slaney_thiebaux_2001} 
%with a \trm{table} and three blocks as objects,
as illustrated in Fig.~\ref{fig:ex-rmdp-bw}.
A state is described using the predicates
$\atm{on}(B,L)$ to denote 
that block $B$ is on top of location $L$;
and
$\atm{goal}(B,L)$ to denote 
{
the agent's task, i.e., a state must be reached in which $B$ is on top of $L$.
}
All actions are based on 
$\atm{move}(B,L)$, denoting the move of block $B$ to some new location $L$. The admissible actions correspond to executable moves, e.g. $A_{s_1} = \{ \atm{move}(2,1), \atm{move}(1,2), \atm{move}(1,table) \}$.
{
For this particular RMDP, the effects of $\atm{move}$ actions are deterministic and 
    cause the insertion and deletion of $\atm{on}$ atoms
    according to the standard blocks world dynamics.
    The $\atm{goal}$ atoms are part of every state and are
    %to be constant across the state space, i.e. not subject to the effects of moves.
    unaffected by moves.
The task of the RMDP is encoded in the reward structure: the reward is $99$ when a state is reached in which all blocks are stacked in ascending order (i.e., $s_2$)
%the last action caused a transition
%to a state in which
%$\atm{on}(B,L)$ holds true for each stated $\atm{goal}(B,L)$,
%Note that partial goal descriptions are allowed.
%Otherwise, the reward is $-1$.
and is $-1$ otherwise.}
The discount rate is $\gamma =1$.
%	Consider a blocks world stacking task, as %illustrated in Fig.~\ref{fig:ex-rmdp-bw}.
%        The signature is defined by $\func = \{ \trm{table}/0, 0/0, 1/0, 2/0 \}$, $\pred_S = \{ \atm{on}/2, \atm{goal}/2 \}$, and $\pred_A = \{ \atm{move}/2 \}$. 

            %For more details, see Section~\ref{sec:blocks}.
            %Relational representations of the states $s_1$ and $s_2$ are given,
            %as well as a 
	   %The state space $S$ is the set of all 13 valid stacking configurations
	   %	as denoted by $\atm{on}/2$.
	   %	The goal description, denoted by $\atm{goal}/2$, stays constant throughout the state space.
	   %The action space $A$ contains all the possible moves, denoted by $\atm{move}/2$.
	   %	The admissibility of moves is restricted to those that make sense for any given configuration.
	   %	For example, $A_{s_1} = \{ \atm{move}(2,1), \atm{move}(1,2), \atm{move}(1,table) \}$.
	   %The transitions are deterministic, defined to be consistent with the usual effects of moves.
 	  %The reward is $99$ if a goal state is reached, $-1$ otherwise, and $\gamma=1$.
	   %For example, performing the action $a_1$ in state $s_1$
	   %yields a new state $s_2$ and a reward of $99$, 
	   %since $s_2$ matches the goal description.
	   %For more on blocks worlds see Section~\ref{sec:blocks}.
	\end{example}

A \emph{policy}
    %$\pi \fdef \Psi \to [0,1]$,
    $\pi(a \mid s)$ defines the probability of taking action $a \in A_s$ in state $s$,
        %models the decision making of the agent
	i.e. $\rv A_t \sim \pi(\ \cdot \mid \rv S_t)$.
The goal is to find an optimal policy $\pi_\star$ that maximises
	the expected return  for all states,
	that is
	$\ex_{p,\pi_\star}\left[\rv G_t \mid \rv S_t = s \, \right] \geq \ex_{p,\pi}\left[\rv G_t \mid \rv S_t = s \, \right]$ %\forall 
    for all $\pi, s, t$. 
    %where $v_\pi$ is the \emph{state-value function} for policy $\pi$ and $v_\star$ is the \emph{optimal state-value function}.
A classic algorithm to find $\pi_\star$ without prior knowledge of $p$ is {\emph \qlearning}~{\citep{Watkins1992}}.
Given a step-size parameter $\alpha \in (0,1]$
%, a \emph{behaviour policy} $\pi$
and an \emph{action-value function} $\rv Q \fdef \Psi \to \reals$,
the version we use  \citep[Chapter~6.5]{sutton_barto_2018}
is
governed by the update
\[
\rv Q(\rv S_t, \rv A_t) \gets \rv Q ( \rv S_t, \rv A_t) + \alpha \cdot \left( \rv R_{t+1} + \gamma \cdot \underset{a}{\max} \bigl \{ \rv Q(\rv S_{t+1}, a) \bigr \} - \rv Q(\rv S_t,\rv A_t) \right).
\]
An approximation of $\pi_\star$ is obtained by selecting actions based on ${\arg \max} \{ \rv Q (s,a) \mid a \in A_s \}$.

\subsection{CARCASS Abstractions}

We now introduce the CARCASS framework~\citep*[Chapter~5]{otterlo_2008_phd},
discussing its syntax, semantics and finally an abstract version of {\qlearning}.
% We start with refining MDPs to relational MDPs.
% Then we define abstract states and actions,
% followed by the CARCASS syntax and semantics,
% and finally an abstract version of {\qlearning}.

% A CARCASS has two aspects: a covering relation $\cov$,
%     defined via SLDNF-resolution,
% and a decision list.
% %, defined
% %by the ordering of the abstract states.
% %A CARCASS is interpreted is a decision list.
% %The intuition behind the semantics is as follows.
% Intuitively, to choose an abstract state-action pair for some $(s,a) \in \Psi$,
% this list can be traversed until the first $\ccs_i$ is found which covers $s$.
% %(i.e. $s \in \cov (\ccs_i)$).
% %and, since it is the first in the list, also
% %$s \in \cc{\ccs_i}$.
% Then, $\ccas_{\ccs_i}$ can be searched for some $\cca_{i,j}$ such that that $(\ccs_i, \cca_{i,j})$ covers $(s,a)$.

%\begin{definition}[Abstract States and Actions {\citep[p.~253]{otterlo_2008_phd}}]
%Let $\Sigma$ be a PL1-Signature and $\var$ a set of variables.
%An \emph{abstract state} $\ccs$ is of the form 
%	$\ccs \defined l_1, \ldots, l_k$, such that $k \geq 1$ and
%		every $l_i$ is a naf-literal over $\Sigma$ and $\var$.\footnote{
%			This includes built-in atoms. Moving away from Prolog, strong negation and aggregates may also be permitted.
%		}
%An \emph{abstract action} $\cca$ is a single predicate atom $\cca \defined \atm{p}(\bar t)$ over $\Sigma$ and $\var$.
%\end{definition}

%\paragraph{CARCASS syntax.}
%\begin{definition}[CARCASS Syntax and Semantics {\citep[p.~253]{otterlo_2008_phd}}]
	%\emph{Syntax.} 
        Let  $\Sigma$
	be a PL1-signature
	and $\var$ a set of variables.
    \begin{itemize}
    \item An \emph{abstract state} $\ccs$ 
    %is of 
    has the form 
        $\ccs \defined l_1, \ldots, l_k$, 
        %such that 
        $k \geq 1$, 
        %where all $l_i$ are
        with all $l_i$ naf-literals over $\Sigma$ and $\var$.
            % \footnote{
            % This includes built-in atoms. Moving away from Prolog, strong negation and aggregates may also be permitted.
            % }
    \item An \emph{abstract action} $\cca$ is a single predicate atom $\cca \defined \atm{p}(\bar u)$ over $\Sigma$ and $\var$.
    \item A \emph{CARCASS} $\abs C = \carcass$
	is a finite %ordered 
    list of $n \geq 1$ elements,
	such that every $\ccs_i$ is an abstract state 
        and every $A_{\ccs_i} = \{ \cca_{i,1}, \ldots, \cca_{i,m_i} \}$ is a set of $m_i \geq 1$ abstract actions
        that are \emph{admissible} in $\ccs_i$.
        The variables occurring in $A_{\ccs_i}$ must also occur in $\ccs_i$.    
 %    $\ccs_1, \ldots, \ccs_n$ are 
	% abstract states and
	% $\abs A_{\ccs_1}, \ldots, \abs A_{\ccs_n}$
	% are sets of abstract actions over $\Sigma$ and $\var$.
	% Every 
	% $A_{\ccs_i} = \{ \cca_{i,1}, \ldots, \cca_{i,m_i} \}$
	% denotes 
	% the set of 
	% abstract actions
	% that are
	% \emph{admissible} 
	% in the abstract state $\ccs_i$.
	% For every $1 \leq i \leq n$ and $1 \leq j \leq m_i$,
	% the variables occurring in $\cca_{i,j}$ must
	% also occur in $\ccs_i$.
	Furthermore, we let
	$\abs S = \{ \ccs_i \mid 1 \leq i \leq n \}$
	and
	$\abs \Psi = \{ (\ccs_i, \cca_{i,j}) \mid  1 \leq i \leq n, 1 \leq j \leq m_i \}$.
        %\emph{Semantics.} 
    \end{itemize}

%\paragraph{CARCASS semantics.}
The semantics of a CARCASS $\abs C$ is defined via a covering relation $\cov$, which uses SLDNF-resolution.
%and a decision list, defined by the order of the elements in $\abs C$.
% %, defined
% %by the ordering of the abstract states.
% %A CARCASS is interpreted is a decision list.
% %The intuition behind the semantics is as follows.
Intuitively, to choose an abstract state-action pair for some concrete pair $(s,a)$,
$\abs C$ is interpreted as a \emph{decision list}  that is  traversed until the first $\ccs_i$ is found which covers $s$.
Then, some $\cca_{i,j}$ with $(\ccs_i, \cca_{i,j})$ covers $(s,a)$ is chosen from $\ccas_{\ccs_i}$.
Formally, given an RMDP $M = \rmdp$
and $\abs C$,
%a CARCASS $\abs C$, 
%    we define 
for every $\ccs_i \in \abs S$
        and $(\ccs_i, \cca_{i,j}) \in \abs \Psi $:
    % Given a set of states $S$ and a set of admissible state-action pairs 
    %     %Now let $M = \rmdp$ be an RMDP of matching signature.
    %     %and 
    %     %let $C = \carcass$ be a CARCASS of matching signature.
    %     For every $\ccs_i \in \abs S$
    %     and $(\ccs_i, \cca_{i,j}) \in \abs \Psi $
    %     we define:
            \begin{align*}
	       \cov (\ccs_i ) &\defined \{ \  s \in S \mid P_s \sldnf \ccs_i \ \}, \\
	       \cov (\ccs_i , \cca_{i,j} ) &\defined \{ \  (s,a) \in \Psi \mid P_s \sldnf \ccs_i \theta \text{ and } a = \cca_{i,j} \theta \ \}, \\
	       \cc{\ccs_i} &\defined \{\  s \in \cov(\ccs_i)\mid s \not \in \cov(\ccs_k) \text{ for all } 1 \leq k < i \ \}, \\
	       \cc{\ccs_i, \cca_{i,j}} &\defined \{\  (s,a) \in \cov(\ccs_i, \cca_{i,j}) \mid s \not \in \cov(\ccs_k) \text{ for all } 1 \leq k < i  \ \}, \\
	       \ccagnd{\ccs_i, \cca_{i,j}}{s} &\defined \{\  a \mid (s,a) \in \cc{\ccs_i, \cca_{i,j}} \ \}.
            \end{align*}
	%\end{definition}
\begin{example}[Blocks World cont'd]\label{ex:cc-bw}
	An example CARCASS for a $3$-blocks world is given in Table~\ref{tab:example-carcass}.
	Intuitively, $\ccs_1$ captures all states with a tower of two blocks,
	$\ccs_2$ captures all states with all blocks on the table,
	and $\ccs_3$ captures all states with a tower of three blocks.
	\end{example}
\begin{table} \small
	\caption{Example CARCASS~\citep[p.~253]{otterlo_2008_phd}}
	\label{tab:example-carcass}
	\centering
	\begin{tabular}{l l}
	\toprule
	$\ccs_1$:	&	$\{ \atm{on}(A,B), \atm{on}(B,\trm{table}), \atm{on}(C, \trm{table}), A \not = B, B \not = C \} $ \\
	$\cca_{1,1}, \cca_{1,2}, \cca_{1,3}$:	&	$\atm{move}(A,C), \atm{move}(C,A), \atm{move}(A,\trm{table})$ \\\\
	$\ccs_2$:	&	$\{ \atm{on}(A,\trm{table}), \atm{on}(B,\trm{table}), \atm{on}(C, \trm{table}), A \not = B, B \not = C, A \not = C \} $ \\
	$\cca_{2,1}, \ldots, \cca_{2,6}$: &	$\atm{move}(A,B), \atm{move}(B,A), \atm{move}(A,C), $ \\
										&	$\atm{move}(C,A), \atm{move}(B,C), \atm{move}(C,B) $ \\\\
	$\ccs_3$: 	&	$\{ \atm{on}(A,B), \atm{on}(B,C), \atm{on}(C, \trm{table}), A \not = B, B \not = C, C \not = \trm{table} \} $ \\
	$\cca_{3,1}$:	&	$\atm{move}(A,\trm{table})$ \\
	\bottomrule
	\end{tabular}
	\end{table}
Algorithm~\ref{alg:ql-carcass}
illustrates
how {\qlearning} can operate on the abstract representations of a CARCASS\@.
The algorithm assumes that the task environment is explored in a series of episodes, each
with a history
%of interactions
$\rv H = (\rv S_0, \rv A_0, \rv R_1, \rv S_1, \ldots )$.
%on the concrete level.
%In addition, 
The CARCASS semantics is used to derive
an abstract series of interactions
$\rva H = (\rva S_0, \rva A_0, \rv R_1, \rva S_1, \ldots )$,
to which {\qlearning} is applied,
working with abstract policies and value functions.
%an abstract policy $\abs \pi (\abs a \mid \abs s)$ and an abstract action-value function $\rva Q (\abs s, \abs a)$.
%action-value function $\rva Q \fdef \abs \Psi \mapsto \reals$.
% Abstract actions are chosen based on an abstract policy $\abs \pi \fdef \abs \Psi \to [0,1]$.
\begin{algorithm}[tb]
    \small
		\KwIn{RMDP $M=\rmdp$, 
		CARCASS $\abs C=\carcass$,\newline
		initial abstract action-value function $\rva Q \fdef \abs \Psi \to \reals$,
            step size $\alpha \in (0,1]$}
            \KwOut{Learned abstract action-value function $\rva Q$ (ideally $\rva Q \approx q_\star$)}
		\ForEach{\emph{episode}} {
            %\Repeat( for every episode ){maximum number of episodes reached}{
			Initialize the starting state $\rv S_0$\;
                Find $ \rva S_0 \in \abs S$ for which $\rv S_0 \in \cc{\rva S_0}$\;
			\Repeat( for every time point $t = 0, 1, 2, \ldots$ ){end of episode \emph{\textbf{or}} maximum number of steps reached }{

                    Choose $\rva A _t \in \abs A_{\rva S_t}$ using an abstract behaviour policy derived from \rva Q\;
                    %Use behavior policy to choose $\rva A_t \in \abs A_{\rva S_t}$.\;
					% \tk check out sutton barto algorithm for inspiration
				%\eIf{exploration}{
				%	Use exploration strategy to choose $\rva A_t \in \abs A_{\rva S_t}$.\;
				%}{
				%	$\rva A_t \gets \arg\max \left\{ \rva Q \left(\rva S_t, \cca \right) \mid \cca \in \abs A_{\rva S_t} \right\} $.
				%} 
				Choose a random action $\rv A_t \in \ccagnd{\rva S_t, \rva A_t}{\rv S_t}$\;
				Take action $\rv A_t$, observe $\rv S_{t+1}$, $\rv R_{t+1}$\;
				Find $ \rva S_{t+1} \in \abs S$ for which $\rv S_{t+1} \in \cc{\rva S_{t+1}}$\;
				$\rva Q\left(\rva S_t, \rva A_t\right) \gets \rva Q\left(\rva S_t, \rva A_t \right) + \alpha \cdot \biggl( \rv R_{t+1} + \gamma \cdot \underset{\cca}{\max} \left\{ \rva Q \left(\rva S_{t+1}, \cca \right) \right\} - \rva Q \left(\rva S_t, \rva A_t \right) \biggr)$\;
				% $\rva Q\left(\rva S_t, \rva A_t\right) \gets \rva Q\left(\rva S_t, \rva A_t \right) + \alpha \cdot \biggl( \rv R_{t+1} + \gamma \cdot \max \left\{ \rva Q \left(\rva S_{t+1}, \cca \right) \Bigm| \cca \in \abs A_{\rva S_{t+1}} \right\} - \rva Q \left(\rva S_t, \rva A_t \right) \biggr)$.\;
                
			}
		}
            \Return \rva Q
 		\caption{{\qlearning} for CARCASSs \citep[p.~258]{otterlo_2008_phd}}
		\label{alg:ql-carcass}
	\end{algorithm}

%\paragraph{Convergence.}
%A number of results establish the \emph{convergence} of abstract {\qlearning} 
%under certain conditions, 
%such as when the behaviour policy is stationary \citep{singh_1994b}
%or 
%if the abstraction maintains certain properties of the concrete MDP,
%like \emph{$q_\star$-irrelevance} for state abstractions \citep{li_walsh_2006}
%or \emph{MDP homomorphisms} \citep{ravindran_2004_phd,ravindran_barto_2003} for state-action pair abstractions.
%Although there are no convergence guarantees in general 
%(and even examples of failure \citep{li_walsh_2006}),
%Sutton and Barto \citep[p.~263]{sutton_barto_2018}
%conjectured that such guarantees exist 
%if the behaviour policy is sufficiently close to the target policy,
%pointing out that they are not aware of
%any observed cases of divergence for this case.

%%%%%%%%%%%%%%%%%%%%%%%%%%%%%%%
\section{ASP-Based Abstractions}\label{sec:asp-carcass}
%%%%%%%%%%%%%%%%%%%%%%%%%%%%%%%

We provide a method
to encode CARCASS abstractions in ASP
and show how %the resulting ASP encoding 
it
can be used in an online learning setting.
We also extend the method 
%such that 
to allow for additional background knowledge.
%can be added.

\subsection{ASP Encoding}
%In the following, 
Let  $\abs C=\carcass$ be a CARCASS
and $M=\rmdp$ an RMDP with a shared signature $\Sigma = \rmdps$.
We assume that none of the function- and predicate names introduced below occur in $\Sigma$.

\begin{definition}[Labelling Function]
	To reason about abstract states and state-action pairs as objects,
	we define a \emph{labelling function} as a one-to-one mapping $\lambda \fdef \abs S \cup \abs \Psi \mapsto \term$
	from abstract states and state-action pairs to a set of ground terms.
		%\item Labelling function $\lambda \fdef \abs S \cup \abs \Psi \to \term $
	\end{definition}
We now present our CARCASS encoding. To this end, we introduce several new atoms: 
%The encoding introduces several new atoms.
	\begin{itemize}
	\item 
$\atm{sIdx}(\lambda(\ccs_i),i)$ states 
%the existence of 
an abstract state $\ccs_i$ (identified by its label)
%            and its 
is at index $i$ in the decision list.

	\item 
$\atm{sCov}(\lambda(\ccs_i),(\bar T))$ states 
%the existence of 
a covering relation between $\ccs_i$ and some state $s$ exists. Denoting all free variables occurring in $\ccs_i$ by $\bar V = V_1, \ldots, V_m$,
% The tuple
$(\bar T)$ represents a ground substitution $\bar V \theta = \bar T$,
reminiscent of the original 
%definition of 
\textit{Cov}, where $P_s \sldnf \ccs_i \theta$.
	\item 
$\atm{sChoice}(\lambda(\ccs_i))$ states $\ccs_i$ as ``chosen'', such that 
        all abstract states covering some concrete state can be enumerated
		as answer sets.
        %(by the choice rule in $P_\textit{Cov}$).
	\item 
$\atm{aCov}(\lambda(\ccs_i, \cca_{i,j}), p(\bar t))$ states a covering relation 
		between $(\ccs_i, \cca_{i,j})$ (identified by its label) and a concrete pair $(s,a)$ exists
		in each answer set where $\ccs_i$ is ``chosen''.
            Note that we use two representations for the covered action $a$: as atom $a = \atm{p}(\bar t)$, as originally defined; and as the term $p(\bar t)$ in \atm{aCov}.
	\end{itemize}
Armed with these auxiliary predicates, we develop the CARCASS encoding by combining several programs that encode different CARCASS components. 

We shall use a program $P_{\ccs_i}$ that encodes a single abstract state, a program $P_{\ccs_i, \cca_{i,j}}$ that encodes the covering relation for abstract state-action pairs,
and programs $P_{Cov}$ and $P_{\hat{C}}$
that translate the decision list semantics
%This can be done by
following the guess-and-check methodology. 
%The guessing part 
Program
$P_{Cov}$ produces solution
candidates 
%for example by
using a choice rule,
and $P_{\hat{C}}$ %is the enhancement with 
% consists of
adds a
constraint to eliminate unwanted %solution 
candidates.
Given a concrete state $s$, represented by program $P_s$, and 
its set $A_s$  of admissible concrete actions, represented by $P_{A_s}$,
$P_s \cup P_{A_s} \cup P_{Cov}$ yields one solution candidate 
per covering abstract state. Furthermore, 
$P_s \cup P_{A_s} \cup P_{\hat{C}}$
eliminates all but one such candidate, corresponding to the selected abstract state via the CARCASS semantics.

%Their intended use is in the context of a concrete state $s$ (represented by program $P_s$) and its set $A_s$  of admissible concrete actions (represented by $P_{A_s}$);
%Recall that $s$ is represented using program $P_s$ and actions $a=p(\bar t)$ as single atoms and the set of admissible actions as program $P_{A_s}$;
%, such that only the intended answer sets remain. 
%For our application, we 
%specifically, we aim
%for $P_s \cup P_{A_s} \cup P_{Cov}$ to produce one solution candidate 
%per %for each c
%covering abstract state and for $P_s \cup P_{A_s} \cup P_{\hat{C}}$
%to eliminate all but one such candidate, corresponding to the selected abstract state via the CARCASS semantics.
\begin{definition}[CARCASS ASP Encoding]\label{def:cc-asp}
	Let $\ccs_i \in\abs S$ be an abstract state
	and let $\bar V = V_1, \ldots, V_m$ denote all the variables occurring in $\ccs_i$. Then, we define programs as follows:
\begin{itemize}
    \item $P_{\ccs_i}$ encodes the CARCASS covering relation for $\ccs_i$ by
    \begin{align*}
		P_{\ccs_i} &\defined \left\{ \begin{array}{l@{~}l@{~}l}
			\atm{sIdx}(\lambda(\ccs_i), i). %&&  \\
		  	\qquad \atm{sCov}(\lambda(\ccs_i), (\bar V)) & \leftarrow & \ccs_i .
			\end{array}\right\}.
    \end{align*}
 %   $P_{\ccs_i}$ single abstract state.
%The formulation makes use of the predicates sIdx, sCov.
The first rule is a fact defining \atm{sIdx} as just described. The head of the second rule
defines \atm{sCov}; its body consists of the literals in $\ccs_i$. The ground instances
of this rule reflect all of the possible ground substitutions for the variables in $\ccs_i$.
    \item $P_{\ccs_i, \cca_{i,j}}$ encodes the CARCASS covering relation for $(\ccs_i, \cca_{i,j}) \in \abs \Psi$,
	with $\cca_{i,j} = \atm{p}(\bar u)$, by
    \begin{align*}
    P_{\ccs_i, \cca_{i,j}} &\defined \left\{ \begin{array}{lll}
		  	\atm{aCov}(\lambda(\ccs_i, \cca_{i,j}), \text{p}(\bar u))
			&\leftarrow&
			\atm{sChoice}(\lambda(\ccs_i)),
			\atm{sCov}(\lambda(\ccs_i), (\bar V)),
	        \atm{p}(\bar u).
			\end{array}\right\},
    \end{align*}
    where $p/n$ is a fresh function name of the same arity as 
     $\atm{p}/n$.
     The head of this rule defines \atm{aCov}; its body reflects the definition of $\cov(\ccs_i, \cca_{i,j} )$, namely the coverage of $\ccs_i \theta$ (by \atm{sCov}) under the possible ground substitutions $\theta$ of the rule (in particular the ground substitutions $\bar V \theta$ of $\bar V$),
     and the admissibility of
     concrete actions $a = \atm{p}( \bar u) \theta$ that are
     ground instances of $\cca_{i,j}$ under said substitutions.
     %Being onl 
     %ensures (in the context of state $s$) 
     %with admissible concrete actions $A_s$
     %that $\ccs_i$ is ``chosen'',
     %(by \atm{sChoice}, see below), 
     %that a covering relation exists between $\ccs_i$ and $s$,
     %(by \atm{sCov}), 
     %and that some concrete action matching $\cca_{i,j}$ is admissible in $s$.
     %, i.e.  $\atm{p}(\bar u)\theta = \cca_{i,j} \theta \in A_s$. 
     %The ground instances of this rule reflect all possible ground substitutions for the variables in $\ccs_i$ (obtained through $\bar V$), a subset of which are the variables in $\cca_{i,j}$. The ground instances of $\cca_{i,j}$ are concrete actions, which may or may not be admissible.%in turn correspond to possibly admissible concrete actions.
    \item $P_\textit{Cov}$ encodes the covering relations for all abstract states and state-action pairs by 
  \begin{align*}
  P_{\textit{Cov}} &\defined  \bigcup_{1 \leq i \leq n} \Big( P_{\ccs_i} \cup \bigcup_{1 \leq j \leq m_i} P_{\ccs_i, \cca_{i,j}} \Big) \cup
			 \Big\{ 1 = \{ \atm{sChoice}(R) : \atm{sCov}(R, \_ ) \ \}. \Big\}.
\end{align*}
%the program $P_{\textit{Cov}}$ contains the definitions of the covering relations for all abstract states. 
It 
%also contains 
includes a choice rule, in which the labels of all covering abstract states are made
available for the choice, but only one can be chosen per answer set.
\item $P_{\abs C}$ encodes the full CARCASS semantics by
\begin{align*}
P_{\abs C} &\defined P_{\textit{Cov}} \cup 
			\left\{ 
			\begin{array}{l}
				\leftarrow  \atm{sChoice}(R1), \atm{sCov}(R2, \_ ), %\\
				\atm{sIdx}(R1,I1), \atm{sIdx}(R2,I2), I2 < I1.
				\end{array}
			\right\}.
\end{align*}      
%For our application, we aim to produce one solution candidate for each covering abstract state. Then, 
\noindent
The constraint added to $ P_{\textit{Cov}}$ eliminates all solution candidates except for the one
corresponding to the minimal covering state with respect to the decision list ordering.

\end{itemize}
  
%	Then, $\ccs_i$ is encoded as $P_{\ccs_i}$.
%   An abstract pair $(\ccs_i, \cca_{i,j}) \in \abs \Psi$ 	with $\cca_{i,j} = \atm{p}(\bar u)$ 	is encoded as $P_{\ccs_i, \cca_{i,j}}$, 	containing a fresh function name $p/n$ with the same arity as 
%    %the predicate name 
%    $\atm{p}/n$.
% The covering relation is encoded by $P_\textit{Cov}$
% and the full CARCASS semantics by $P_{\abs C}$.
%	\begin{align*}
	%	P_{\ccs_i} &\defined \left\{ \begin{array}{lll}
	%		\atm{sIdx}(\lambda(\ccs_i), i). %&&  \\
	%  	   \qquad \atm{sCov}(\lambda(\ccs_i), (\bar V)) & \leftarrow & \ccs_i .
	%	\end{array}\right\}\\
		%P_{\ccs_i, \cca_{i,j}} &\defined \Big\{ \begin{array}{lll}
	%  	\atm{aCov}(\lambda(\ccs_i, \cca_{i,j}), \text{p}(\bar u))
		%\leftarrow&
		%	\atm{sChoice}(\lambda(\ccs_i)),
		%  \atm{sCov}(\lambda(\ccs_i), (\bar V)),
	    %    \atm{p}(\bar u).
		% \end{array}\Big\}\\
		%P_{\textit{Cov}} &\defined  \bigcup_{1 \leq i \leq n} \left( P_{\ccs_i} \cup \bigcup_{1 \leq j \leq m_i} P_{\ccs_i, \cca_{i,j}} \right) \cup
			% \Big\{ 1 = \{ \atm{sChoice}(R) : \atm{sCov}(R, \_ ) \ \}. \Big\} \\
		%P_{\abs C} &\defined P_{\textit{Cov}} \cup 
		% \left\{ 
		%\begin{array}{lll}
		%	&\leftarrow & \atm{sChoice}(R1), \atm{sCov}(R2, \_ ), \\
		%	&&\atm{sIdx}(R1,I1), \atm{sIdx}(R2,I2), I2 < I1.
			%	\end{array}
		%	\right\}
	%\end{align*}
\end{definition}

%In more detail, the programs work as follows.
%TODO: Describe the programs.
%$P_{\ccs_i, \cca_{i,j}}$ encodes the covering relation for a single abstract state-action pair.

%Next, we translate the decision list characteristic. This can be done by following the guess-and-check methodology The guessing part produces solution candidates, for example by using a choice rule. The checking part usually consists of constraints which eliminate solution candidates, such that only the intended answer sets remain. For our application, we aim to produce one solution candidate for each covering abstract state. Then, a constraint eliminates all solution candidates except for the one corresponding to the minimal covering state with respect to the decision list ordering.

%For the guessing part, the program $P_{\textit{Cov}}$ contains the definitions of the covering relations for all abstract states. It also contains a choice rule, in which the labels of all covering abstract states are made available for the choice, but only one can be chosen per answer set.

%The program $P_{\abs C}$ adds a constraint eliminating all interpretations in which the index of he chosen covering abstract state is not minimal. This is the checking part of the program.

\begin{example}[Blocks World cont'd]\label{ex:cc-bw-asp}
	To illustrate the ASP encoding, consider the CARCASS from Example~\ref{ex:cc-bw}, in particular the abstract state $\ccs_2$ and the abstract pair $(\ccs_2, \cca_{2,1})$.
	We choose descriptive names as labels, e.g. 
	$\lambda(\ccs_2) = \texttt{"3towers"}$ and
	$\lambda((\ccs_2, \cca_{2,1})) = \texttt{"3t-mAB"}$.
	Then:
	\[
		P_{\ccs_2} = \left\{ \begin{array}{lll}
			\atm{sIdx}(\texttt{"3towers"}, 2).&&  \\
		  	\atm{sCov}(\texttt{"3towers"}, (A,B,C)) & \leftarrow &\atm{on}(A,\trm{table}), \atm{on}(B,\trm{table}), \atm{on}(C, \trm{table}), \\
			&& A \not = B, B \not = C, A \not = C.
			\end{array}\right\},
	\]
	\[
		P_{\ccs_2, \cca_{2,1}} = \left\{ \begin{array}{l@{~}l}
		  	\atm{aCov}(\texttt{"3t-mAB"}, \trm{move}(A,B))
			\leftarrow &\ \atm{sChoice}(\texttt{"3towers"}),\\
			%\qquad\qquad\qquad\qquad\qquad\quad
            &\ \atm{sCov}(\texttt{"3towers"}, (A,B,C)), %\\
	          \atm{move}(A,B).
			\end{array}\right\}.
	\]
        The encodings for the other abstract states $\ccs_1$ and $\ccs_3$, and their corresponding abstract state-action pairs follow the same pattern.
        The encoding for $P_{\textit{Cov}}$ is the union of the ones for all abstract states and state-action pairs with the choice rule added as defined above, i.e. $P_{\textit{Cov}} = (P_{\ccs_1} \cup P_{\ccs_1, \cca_{1,1}} \cup P_{\ccs_1, \cca_{1,2}} \cup P_{\ccs_1, \cca_{1,3}}) \cup (P_{\ccs_2} \cup  P_{\ccs_2, \cca_{2,1}}  \cup   \ldots \cup P_{\ccs_2, \cca_{2,6}} )\cup ( P_{\ccs_3} \cup P_{\ccs_3, \cca_{3,1}}) \cup
			 \left\{ 1 = \{ \atm{sChoice}(R) : \atm{sCov}(R, \_ ) \ \}. \right\}$.
        Finally, $P_{\abs C}$ is defined as above. The complete encoding  is available in the supplementary material. %\ref{sec:enc:example}.
	\end{example}
%The logic programming aficionado may ask why we use a guess-and-check %approach and not, e.g.,
%%Could also be done using 
%%employ 
%aggregates. %or 
%%via 
%%$not\ {-}optimal$ (as for the Prolog encoding).
%This is because guess-and-check gives us more flexibility in how we %combine the programs to model different aspects of the CARCASS. %Assuming we have, e.g.,  a program $P_\Psi$ that generates all state-%action pairs as answer sets, then our encoding enables us to express %$\cov (\ccs_i)$ as 
%%$\as( P_\Psi \cup P_\cov \cup \{ \gets \naf \atm{sChoice} (\lambda %(\ccs_i)) \}$.
%the answer sets of $P_\Psi \cup P_\cov \cup \{ \gets \naf %\atm{sChoice} (\lambda (\ccs_i))$, where $P_\Psi$, $P_\cov$ are %defined using guess and check, as above.

\subsection{Correctness}
%We can establish the following proposition (proof in the appendix): 
As for the correctness of the encoding, we establish the following result (proof in the supplementary material).% \ref{proof:correctness}). 
	\begin{proposition}\label{prop:correctness}
            \input{prop-correctness}
	\end{proposition}
	%\begin{proof}
	%See Appendix~\ref{proof:correctness}.
	%\end{proof}

\noindent
%The following properties (P1)-(P3) 
%follow:
%of the translation are then entailed. 
Based on this result, desired properties of the encoding can be concluded. 
\begin{corollary} For $M$, ${\abs C}$, $\lambda$, $s$, and ${\abs s}$ as in Prop.~\ref{prop:correctness}, the following properties (P1)-(P3) hold:
	\begin{enumerate}[leftmargin=*,label={(P\arabic*)},itemsep=2pt,topsep=2pt]
%\begin{itemize}[leftmargin=*,label=,itemsep=1pt]
	\item
    %The answer sets 
    $\as (P_s \cup P_{A_s} \cup P_\textit{Cov})$
	corresponds one-to-one 
    with the abstract states that cover $s \in S$.
	\item % [(P2)]
    %Each such answer set, 
    Each $I \in \as (P_s \cup P_{A_s} \cup P_\textit{Cov})$,  corresponding with the chosen abstract state $\ccs_i$,
             contains a listing of the admissible abstract actions $\cca_{i,j} \in \abs A_{\ccs_i}$
	    and their covered admissible concrete actions $\atm{p}(\bar t) = a \in A_s$
            in the form of $\atm{aCov}(\lambda(\ccs_i, \cca_{i,j}), \text{p}(\bar t))$ atoms.
	\item %[(P3)] 
    $\as (P_s \cup P_{A_s} \cup P_{\abs C})$ contains
	at most one such answer set, corresponding to the covering abstract state
	with the lowest index.
%\end{itemize}
\end{enumerate}
\end{corollary}
%\noindent
Property (P1) follows from the choice rule in $P_\textit{Cov}$.
To verify (P2), consider  the rule
\[
	\atm{aCov}(\lambda(\ccs_i, \cca_{i,j}), \text{p}(\bar u))
	\leftarrow
	\atm{sChoice}(\lambda(\ccs_i)),
	\atm{sCov}(\lambda(\ccs_i), (\bar V)),
	\atm{p}(\bar u).
\]
in $P_{\ccs_i, \cca_{i,j}}$, which has a ground instance for every occurrence of
$\atm{sCov}$ in $I \in \as( P_s \cup P_{A_s} \cup P_\textit{Cov} )$.
The body of this ground instance will be true if
(i) $\ccs_i$ is ``chosen'' in $I$ (by the choice rule in $P_\textit{Cov}$), (ii)
the substitution $\theta$ related with the ground instance is such that $P_s \sldnf \ccs_i \theta$,
and (iii)
there exists an admissible concrete action $\atm{p}(\bar u)\theta =a \in A_s$.
%(i.e. is a fact in $P_{A_s}$).
These criteria correspond with the ones in 
the definition of $\cov(\ccs_i, \cca_{i,j})$
and the rule head rightly asserts that $(\ccs_i, \cca_{i,j})$ covers $(s,a)$.
Generalising this argument,
we can see how $P_{\ccs_i, \cca_{i,j}}$ identifies all admissible actions for which the covering relation holds.
To verify (P3),
observe that
$P_{\abs C}$ is just $P_\textit{Cov}$ %with a constraint added,
plus a constraint, 
eliminating all answer sets except %for 
the one where the index of the chosen covering abstract state is minimal.

\subsection{Online Learning for ASP-Based CARCASS Abstractions}

Algorithm~\ref{alg:carcass-asp} illustrates 
	how the ASP-encoding of a CARCASS can be 
	used in an online learning setting such as Algorithm~\ref{alg:ql-carcass}. 
	The ASP encoding used here includes (i) the  programs  discussed  so-far,
	(ii) an optional program $P_B$ with background knowledge,
	and (iii) the program 
	\[ P_{\ccs_\infty} \defined \left\{\ % \begin{array}{l}
					\atm{cState}(\texttt{"true"}, \# sup).		%		\\
				\quad	\atm{cStateCovers}(\texttt{"true"}, ()).
			%\end{array}
            \ \right\}
	\]
        (where $\# sup$ stands for the \emph{supremum}, or $+\infty$)
	representing an empty abstract state $\ccs_\infty = \top$ that trivially covers every concrete state and always has the highest index.
	The admissible actions are %defined as 
    $\abs A_{\ccs_\infty} \defined \{ \cca_{\infty, 1} \}$,
	such that $\ccagnd{\ccs_\infty ,  \cca_{\infty, 1}}{s} \defined A_s$ covers all admissible concrete actions.
\begin{algorithm}[t]
        \small
		\KwIn{Current state $s$ with admissible actions $A_s$, 
                background knowledge $P_B$, and \newline
                %}
		%\KwIn{
            a CARCASS $\abs C=\carcass$ with labels $\lambda$ and encoding $P_{\abs C}$
            %of a }
		%\KwIn{Labelling function $\lambda \fdef \abs S \cup \abs \Psi \to \term$}
		%\KwIn{
            }
            \KwOut{Chosen abstract state $\ccs$, abstract actions $\abs A'_{\ccs} \subseteq \abs A_ {\ccs}$, and $\ccagnd{\ccs, \cca}{s}$ for all $\cca \in \abs A'_{\ccs}$.}
		$ I \gets \text{ An arbitrarily chosen optimal answer set } I \in \mathcal {AS} (P_{\abs C} \cup P_s \cup P_{A_s} \cup P_{\ccs_\infty} \cup P_B)$ \;
		$\ccs \gets \lambda^{-1}(l) \text{ for the only } \atm{sChoice}(l) \in I$ \;
		\eIf{$\ccs = \ccs_{\infty}$}{
			$\hat A' _\ccs \gets \{ \cca_{\infty, 1} \}$ \;
			$\ccagnd{\ccs,  \cca_{\infty, 1}}{s} \gets A_s$ \;
		}{
			$\abs A '_\ccs \gets \left\{ \cca \mid \atm{aCov}(l, p(\bar t)) \in I \land \lambda(\ccs, \cca) = l \right\}$\;
			\For{ $\cca \in \abs A '_\ccs $ } {
				$\ccagnd{\ccs, \cca}{s} \gets \left\{ \atm{p}(\bar t) \mid \atm{aCov}(l, \trm{p}(\bar t)) \in I \land \lambda(\ccs, \cca) = l \right\}$\;
			}
		}
        \Return{$\ccs$, $\abs A'_{\ccs}$, $\left\{ \ccagnd{\ccs, \cca}{s} \right \}_{\cca \in \abs A'_{\ccs}}$}
	\caption{Online interaction for ASP-encoded CARCASSs}
	\label{alg:carcass-asp}
	\end{algorithm}
%\paragraph{Background knowledge.}
As for (ii), 
%The algorithm allows that 
the CARCASS encoding 
can be combined with %additional 
background knowledge
%in the form of an ASP program $P_B$, 
%including the possibility to add 
where 
%the ASP program 
$P_B$ may include weak constraints.
However, if $P_B$ 
%causes the generation of 
leads to multiple (optimal) answer sets,
one needs to be careful that they agree on the chosen abstract state.
%A simple way to ensure this is to add 
This can be ensured by adding the weak constraint 
$
	:\sim \atm{sChoice}(R), \atm{sIdx}(R,I).\; [I@1,R]
$
%to act as a tie breaker.
for tie-breaking.
%on the lowest priority level,provided that other weak constraints are defined only on priority levels 2 or higher.

In line~1 of Alg.~\ref{alg:carcass-asp}, the augmented ASP encoding is handed to an ASP solver, which is expected to return 
	an optimal answer set. 
	If $P_B = \emptyset$, 
    %there exists only
    a single answer set corresponding to the covering abstract state 
	with the smallest index exists, as previously described.
	%If there are 
    In case of multiple optimal answer sets, 
    %the choice of which to return is left to the solver.
    the solver chooses one to return.
	Lines~3--5 are then concerned with the special case in which $\ccs_\infty$ was chosen,
	and
	lines~7--10 deal with the extraction of information from the returned answer set 
    %in case 
    if some other $\ccs \in \abs S $ was chosen.
%\paragraph{Outputs.}
The algorithm returns
%\begin{itemize}
%\item 
the chosen abstract state $\ccs$,
%\item 
a set of abstract actions $\abs A'_{\ccs} \subseteq \abs A_ {\ccs}$ that cover at least one admissible concrete action---to be used instead of $\abs A_{\ccs}$ in the definition of abstract policies---%
	%to avoid choosing abstract actions covering no concrete actions,
%\item a set of abstract actions 
%	$\abs A'_\ccs = \left\{ \cca \in \ccas_\ccs \mid \ccagnd{\ccs, \cca}{s} \not = \emptyset \right\}$ 
%	that cover at least one concrete action 
%\item 
and the sets $\ccagnd{\ccs, \cca}{s}$ for all $\cca \in \abs A'_{\ccs}$.
%\end{itemize}

	%	\paragraph{Admissible actions.}
	%	In our im
	%	$\abs A'_\ccs \subseteq \abs A_ \ccs$
	%	
	%	A peculiarity of both Algorithm~\ref{alg:online-carcass-asp} and Algorithm~\ref{alg:online-carcass-asp-ext}
	%		is the treatment of an abstract action $\cca$ not covering any concrete actions.
	%	
	%		In this case, no related instances of $\atm{cActionCovers}$ are part of the answer set
	%		and $\cca$ will not count as admissible in our implementation.
	%		So, it is possible for one abstract state $\ccs$ to have different sets of admissible actions
	%		in different concrete states.
	%		In our implementation, this is handled by adding the set of admissible actions to the state description,
	%		forming a refined abstract state $\abs x_t  = (\ccs_t, \ccas_{\ccs_t})$.
	%		This way, if at any time points $t_1,t_2$ during the learning process,
	%		we encounter the same CARCASS state ($\ccs_{t_1} = \ccs_{t_2}$)
	%		but their sets of admissible actions are decided to be different ($\ccas_{\ccs_{t_1}} \not = \ccas_{\ccs_{t_2}}$),
	%		we conclude that they are different abstract states $(\abs x_{t_1} \not = \abs x_{t_2})$ for the purposes of online learning.
	%		This is formalised in a new $q$-update, which is amended in Algorithm~\ref{alg:ql-carcass}.
	%		\begin{align*}
	%			\abs q(\abs x_t,\cca_t) \leftarrow \abs q(\abs x_t,\cca_t) + \alpha [ r_{t+1} + \gamma \underset{\cca' \in \abs A_{\ccs_{t+1}}}{\max} \{ \abs q(\abs x_{t+1}, \cca') \} - \abs q(\abs x_t,\cca_t) ]
	%		\end{align*}

%%%%%%%%%%%%%%%%%%%%%%%%%%%%%%%%%%%%%%%%
\section{Case Study: Blocks World}\label{sec:blocks}
%%%%%%%%%%%%%%%%%%%%%%%%%%%%%%%%%%%%%%%%
As a first application, we discuss 
$n$-blocks world stacking tasks
and present the encoding of a state-action pair abstraction, inspired by the
algorithms GN1 and GN2~\citep{slaney_thiebaux_2001},
	%\citep[pp.~129--131]{slaney_thiebaux_2001} citing \citep[pp.~229--230]{gupta_nau_1992}).
	%in which 
    where the size of the abstract state-action space is constant in %the number of blocks.
    $n$.
	As a trade-off, the abstraction does not preserve the optimal policy for the (NP-hard) blocks world planning problem.
We only glance over some of the details here; 
%but note that
	the full CARCASS encoding
    is available in 
    the supplementary material.
    %see \ref{sec:enc:bw}.

%	In the following we discuss our RMDP representation of a blocks world planning task,
%	the background knowledge which we used in combination with the abstraction,
%	and the encoding of the abstraction.

%\paragraph{Blocks World as RMDP.}
%The details of a $3$-blocks world RMDP are given above, in  Example~\ref{ex:rmdp-bw}. 
{
The abstraction was designed with a generalised version of the RMDP from Example~\ref{ex:rmdp-bw} in mind.
We assume an $n$-blocks world RMDP with an arbitrary number of $n \geq 1$ blocks and an arbitrary stacking task, encoded in the reward structure and communicated to the agent via the set of $\atm{goal}$ atoms in the state descriptions.
% Partial stacking tasks, in which the locations of some blocks are not relevant, are also possible.
}
%A state is described using the predicates
%$\atm{on}(B,L)$ to denote 
%that block $B$ is on top of location $L$;
%and
%$\atm{goal}(B,L)$ to denote 
%that $B$ must be on top of $L$ in a goal state.
%All actions are based on 
%$\atm{move}(B,L)$, denoting the move of block $B$ to %some new location $L$.
%The effects of moves are deterministic and according %to the standard blocks world dynamics.
%The reward is $99$ when the last action caused a %transition
%to a goal state, i.e.\
%$\atm{on}(B,L)$ holds true for each stated $\atm{goal}%(B,L)$.
%Note that partial goal descriptions are allowed.
%Otherwise, the reward is $-1$.
%The discount rate is $\gamma =1$.

%\paragraph{Background knowledge.}
Several predicates were defined to support the abstraction,
including the following:
%\begin{itemize}
%\item 
$\atm{clear}(L)$ denotes that location $L$ is clear, 
	i.e., no blocks are located on top, or 
    %if the location 
    $L$ is the table;
%\item 
$\atm{above}/2$ denotes the transitive closure of $\atm{on}/2$; and
%\item 
$\atm{incomplete}(L,B)$ denotes the existence of an \emph{incomplete tower} in its final position, 
%The first argument represents the location of that tower and
%the second argument represents its highest block, with all blocks below or in between being in their final positions.
with location $L$, topmost block $B$, and
other misplaced blocks allowed to occur on top of $B$;
incompleteness means that 
%there exists another block that needs to be added to the tower
some further block(s) must be added
{on top of $B$ to fulfil the stacking task. Note that the definition of $\atm{incomplete}$ depends on both $\atm{on}$ and $\atm{goal}$.}
%\end{itemize}

\begin{example}[Example~\ref{ex:rmdp-bw} cont'd]
We augment $s_1$ 
with the facts 
$\atm{clear}(\trm{table})$,
$\atm{clear}(1)$,
$\atm{clear}(2)$,
$\atm{above}(0,\trm{table})$,
$\atm{above}(1,0)$, 
$\atm{above}(1,\trm{table})$, 
$\atm{above}(2,\trm{table})$, and
 $\atm{incomplete}(\trm{table},1)$.
% with block $2$ still missing to complete the tower.
%\end{example}
%\begin{example}
To further illustrate incomplete towers, consider
{
a different $5$-blocks world RMDP with a (partial) stacking task where blocks $0$--$3$ need to form a tower in ascending order and the position of block $4$ is irrelevant. The corresponding state
}
\[
s_3 = \left\{%\begin{array}{l}
\atm{on}(4,\trm{table}), 
\atm{on}(0,4), 
\atm{on}(1,0), 
\atm{on}(2,1), 
\atm{on}(3,\trm{table}), %\\
\atm{goal}(1,0), 
\atm{goal}(2,1), 
\atm{goal}(3,2) \\
%\end{array}
\right\}
\]
{ is augmented with $\atm{incomplete}(0,2)$, reflecting the fact that blocks $0$, $1$, $2$, and $4$ are in their final position, but $3$ must still be added on top of $2$.
%with $3$ being the next missing block. 
%This is a valid goal tower as block $4$ does not need to be moved to solve the task. 
% Note there is no occurrence of block $4$ in the $\atm{goal}$-atoms of the state description.
For state $s_3 \cup \{ \atm{goal}(4,3) \}$ in yet another $5$-blocks world RMDP with the task to stack all five blocks in ascending order, no goal tower exists since no blocks are in their final position.
}
\end{example}

%\paragraph{CARCASS encoding.}
There are six abstract states, covering the following cases:
	\begin{itemize}[leftmargin=*]
	\item $\ccs_0$: an incomplete tower exists that is clear and the next missing block is clear;
	%\tk Actions:
		%	There are two ways to move
		%	BadTop. First, it can always be moved to the table, a move covered by the abstract
		%	action move(badTop,table). Second, if another tower exists (which may or may not
		%	be goal relevant, incomplete and/or in its final position), the block can be moved on
		%	top of that other tower, as covered by move(badTop,other).14 All other actions are
		%	covered by gutterAction.
	\item $\ccs_1$: an incomplete tower exists but has other misplaced blocks on top;
	\item $\ccs_2$: an incomplete tower that is clear exists, but the next missing block is not clear;
	\item $\ccs_3$: no incomplete tower exists but the first block to construct one is not clear;
	\item $\ccs_4$: no incomplete tower exists and the first block to construct one is clear; and
	\item $\ccs_\infty$: all other states, i.e. goal states.
	\end{itemize}
To illustrate our encoding, we discuss $\ccs_0$ and $\ccs_1$ in detail,
	including their admissible abstract actions.
The abstract state $\ccs_0$ is encoded as
	\[
		P_{\ccs_0} = \left\{ \begin{array}{lll}
			\atm{sIdx}(\trm{r0}, 0).&&  \\
		  	\atm{sCov}(\trm{r0}, (T, N)) 
				& \leftarrow & \atm{incomplete}(\_, T), \atm{clear}(T), \atm{goal}(N, T), \atm{clear}(N).
				%& \leftarrow &\atm{on}(A,\text{table}), \atm{on}(B,\text{table}), \atm{on}(C, \text{table}), \\
			%&& A \not = B, B \not = C, A \not = C.
			\end{array}\right\}.
	\]
The next move towards { fulfilling the stacking task} is to place the next missing block $N$
on top of the incomplete tower $T$, which is captured by the abstract action $\cca_{0,0}$, encoded as
	\[
		P_{\ccs_0, \cca_{0,0}} = \left\{ \begin{array}{lll}
		  	\atm{aCov}(\trm{move}(n,t), \trm{move}(N,T))
			&\leftarrow& \atm{sChoice}(\trm{r0}), \atm{sCov}(\trm{r0}, (T,N)).
			%&\leftarrow& \atm{sChoice}(\texttt{"alltable"}),\\
			%&&\atm{sCov}(\texttt{"alltable"}, (A,B,C)), \\
	        %&& \atm{move}(A,B).
			\end{array}\right\}.
	\]
%As a second abstract action, we 
We also define $\cca_{i,\infty}$
as a catch-all abstract action, covering actions 
%that are 
not covered by $\cca_{0,0}$,
%It is encoded as
encoded as
	\[
		P_{\ccs_i, \cca_{i,\infty}} = \left\{ \begin{array}{l}
		  	\atm{aCov}(\trm{others}, \trm{move}(X,Y))
			\leftarrow \atm{clear}(X), \atm{clear}(Y), X \neq Y, X \neq \trm{table}, \\
			  \qquad\qquad\qquad\qquad\naf \atm{aCov}(move(\_,\_), move(X,Y)), %\\
			%&& 
            \naf \atm{on}(X,Y).
			%&\leftarrow& \atm{incomplete}(\_, T), \atm{clear}(T), \\
			%&& \atm{goal}(N, T), \atm{clear}(N).
			%&\leftarrow& \atm{sChoice}(\texttt{"alltable"}),\\
			%&&\atm{sCov}(\texttt{"alltable"}, (A,B,C)), \\
	        %&& \atm{move}(A,B).
			\end{array}\right\},
	\]
such that it can be reused also for the other abstract states.
Next,  $\ccs_1$ is encoded as
	\[
		P_{\ccs_1} = \left\{ \begin{array}{lll}
			\atm{sIdx}(\trm{r1}, 1).&&  \\
		  	\atm{sCov}(\trm{r1}, (L,T,B)) 
				& \leftarrow & \atm{incomplete}(L, T),
                               \naf \atm{clear}(T),
                               \atm{above}(B, T),
                               \atm{clear}(B).
				%& \leftarrow &\atm{on}(A,\text{table}), \atm{on}(B,\text{table}), \atm{on}(C, \text{table}), \\
			%&& A \not = B, B \not = C, A \not = C.
			\end{array}\right\}
	\]
with the admissible abstract actions $\cca_{1,0}$ and $\cca_{1,1}$:
\begin{align*}
P_{\ccs_1, \cca_{1,0}} &= \left\{ \begin{array}{lll}
		  	\atm{aCov}(\trm{move}(b,\trm{table}), \trm{move}(B,\trm{table}))
			&\leftarrow& \atm{sChoice}(\trm{r1}), %\\
			%&& 
            \atm{sCov}(\trm{r1}, (\_, \_,B)).
			%&\leftarrow& \atm{sChoice}(\texttt{"alltable"}),\\
			%&&\atm{sCov}(\texttt{"alltable"}, (A,B,C)), \\
	        %&& \atm{move}(A,B).
			\end{array}\right\}\\
P_{\ccs_1, \cca_{1,1}} &= \left\{ \begin{array}{lll}
		  	\atm{aCov}(\trm{move}(b,o), \trm{move}(B,O))
			&\leftarrow&
			 \atm{sChoice}(\trm{r1}),
			 \atm{sCov}(\trm{r1}, (L,T,B)), \\
			&& \atm{clear}(O), O \not= B, O \not= \trm{table}.
			%&\leftarrow& \atm{sChoice}(\texttt{"alltable"}),\\
			%&&\atm{sCov}(\texttt{"alltable"}, (A,B,C)), \\
	        %&& \atm{move}(A,B).
			\end{array}\right\}
\end{align*}
To advance $\ccs_1$ towards { fulfilling the stacking task},
all blocks on top of $T$ must be removed.
%starting with 
The topmost block $B$
can be placed either on the table ($\cca_{1,0}$) or 
on top of some other free block ($\cca_{1,1}$).
All other actions are again covered by $\cca_{i,\infty}$.

%\paragraph{Differences to the canonical CARCASS encoding.}
%Note that the abstraction does not strictly adhere
%to Definition~\ref{def:cc-asp}.
%For example, the admissible action atoms were omitted in the rule
%bodies of $P_{\ccs_i, \cca_{i,j}}$,
%since the admissibility of actions is clear from the context.
%Also new is the catch-all action $\cca_{i,\infty}$.
{
%\paragraph{Analogy to GN1 and GN2 \citep{slaney_thiebaux_2001}.}
The intuition behind this abstraction relates back to the algorithms GN1 and GN2 \citep{slaney_thiebaux_2001}.
The abstract state $\ccs_0$ covers case (2) of the GN1 algorithm
and moves covered by $\cca_{0,0}$ are constructive moves.
The abstract states $\ccs_1$ and $\ccs_2$ cover
states that are deadlocked and relate to the two cases in the definition
of the function $\delta_\pi$.
}
%The abstract states 
%$\ccs_3$ and $\ccs_4$ are not immediately applicable
%as they are needed only for stacking tasks with a partial goal description.

%%%%%%%%%%%%%%%%%%%%%%%%%%%%%%%%%%%%%%%%
\section{Case Study: MiniGrid}\label{sec:minigrid}
%%%%%%%%%%%%%%%%%%%%%%%%%%%%%%%%%%%%%%%%

\begin{figure}[t]
		\begin{center}
		\begin{minipage}{0.29\textwidth}
		\centering
		\frame{\includegraphics[height=2cm]{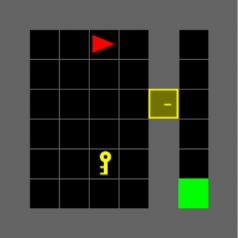}}
		\end{minipage}
		%\hspace{1em}
		\begin{minipage}{0.39\textwidth}
	    \scriptsize
			$	
			s_1  = \left\{ \begin{array}{l}
				\atm{obj}(\trm{goal},(6,6)) \\
				\atm{obj}(\trm{door}(\trm{yellow},\trm{locked}),(5,3)) \\
				\atm{obj}(\trm{key}(\trm{yellow}),(3,5)) \\
				\atm{obj}(\trm{agent}(\trm{east}), (3,1)) \\
				\atm{obj}(\trm{wall}(\trm{grey}), (0,0)) \\
				\atm{obj}(\trm{wall}(\trm{grey}), (0,1)) \\
				\ldots
				\end{array}\right\}
			$
		\end{minipage}
		\begin{minipage}{0.29\textwidth}
		\centering
		\frame{\includegraphics[height=2cm]{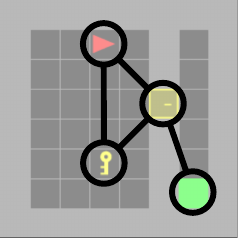}}
		\end{minipage}
		\end{center}
		\caption{Example Minigrid state: 
        %from the Door Key $8 \times 8$ task environment.
		The left image shows a state,
		its relational representation is in the center, and 
		a graph of the room layout, inferred using ASP, is to the right.}
		\label{fig:ex-rmdp-mg}
		\end{figure}

Next, we consider environments
	from the MiniGrid simulation library \citep{boisvert_minigrid_2023}.
	Solving such an environment typically requires the agent
	to navigate through a fixed layout of connected, rectangular rooms to reach the
	location of a goal tile as fast as possible,
	solving simple puzzles and avoiding hazards along the way.
\begin{example}[Door Key Environment]
        In state $s_1$ (Fig~\ref{fig:ex-rmdp-mg}),
	%Consider the  example state from .
	the agent and a key are located in one room 
        and the goal tile in the other, separated by a locked door. 
	%To solve the task, the agent must pick up the key and unlock the door.
    The room layout changes between episodes, which makes the learning task more difficult.
\end{example}
We present the encoding of a state abstraction that is applicable
to a variety of such environments. 
Background knowledge is used 
to infer a room layout from the raw state description,
to build a graph based on the location of objects,
%in that layout, % (e.g. Fig.~\ref{fig:ex-rmdp-mg}),
and to compute the shortest path on that graph from the agent to the goal tile,
representing an ordered set of subtasks that the agent must accomplish.
%It is easy to further constrain 
Paths can be further constrained,
for example to ensure that the key is visited before the door. 
The abstract state is constructed such that
it contains only information relevant to the first node on the path. 
The full encoding of the abstraction is available in the supplementary material. %\ref{sec:enc:mg}.

%\paragraph{MiniGrid as RMDP.}
We assume a MiniGrid configuration in which the current state is fully observable, resembling a birds-eye view as in Fig~\ref{fig:ex-rmdp-mg}, and replace the default state description (a matrix $\mathbf{A} \in \natz ^ { w \cdot h \cdot 3}$, where $w \times h$ is the size of the grid world)
%By default, a MiniGrid state is only partially observable and represented as a numeric matrix.
	%	\footnote{
	%		Some environments also require a \emph{mission space string} as part of
	%		their state description but this is ignored for the purposes of this research.
	%	}
%We apply the \emph{Fully Obs Wrapper},
%such that the state description resembles a birds-eye view of the %grid world,
%and 
with a 
%custom 
relational state description, using the following atoms:
%\vspace{-1ex}
%\begin{itemize}
%\item 
%$\atm{obj}/2$ denotes that some object (first argument) is placed at some grid coordinate (second argument);
	%Objects are represented as functional terms, 
	%with the function name being the name of the object and nested terms describing the object's attributes.
%\item 
$\atm{obj}(O,(X,Y))$ asserts that object $O$ is located at the grid coordinates $(X, Y)$;
%\item 
%$\atm{carries}/1$ denotes that the agent carries some object;
$\atm{carries}(O)$ asserts that the agent carries object $O$;
and
%\item 
$\atm{terminal}$ denotes the end of an episode.
%\end{itemize}
The actions are 
%denoted by% the atoms
	$\atm{right}$,
	$\atm{left}$,
	$\atm{forward}$,
	$\atm{pickup}$,
	$\atm{drop}$,
	$\atm{toggle}$, and
        $\atm{done}$.
	%The following actions are available exist.
		%	%\vspace{-1ex}
		%	\begin{itemize}
		%	\item $\atm{right}$ and $\atm{left}$ causes the orientation of the agent to change clockwise and counter-clockwise, respectively.
		%	\item $\atm{forward}$ causes the agent to move one step forward into the grid cell that is immediately in front of the agent.
		%	\item $\atm{pickup}$ causes the agent to pick up an appropriate object that is located in the grid cell directly in front of the agent.
		%		Only one object can be carried at any given point in time.
		%	\item $\atm{drop}$ causes the agent to drop an object that is currently carried.
		%	\item $\atm{toggle}$ causes the agent to interact with an object in the grid cell directly in front of the agent.
		%		For example, doors can be unlocked (if an appropriate key is currently carried), opened or closed using this action.
		%	\end{itemize}
		%	For environments of this family, all actions are admissible in all states,
		%		even if they make no sense (e.g. pickup when nothing is in front of the agent; moving against a wall)
		%	Actions do nothing if they fail.
	%	Positive rewards are given according to the equation $1 - 0.9 \cdot \frac{t}{t_\text{max}}$ 
	%	if the goal tile was reached with the last action
	%	and within some time limit $t_\text{max}$.
	%	Otherwise, rewards are zero.
	%	%	Note that the time $t$ is not part of the agent's state description,
	%	%	which technically makes the environment partially observable.
    
%\paragraph{Background knowledge.}
The key components of our abstraction, inferred using background knowledge,
are represented with the following atoms:
%\begin{itemize}
%\item 
$\atm{oriented}(O)$ asserts the orientation of the agent;
%($\trm{north}$, $\trm{south}$, $\trm{east}$, or $\trm{west}$).
%\item 
$\atm{fronting}((X,Y))$ 
%asserts 
that $(X,Y)$ is directly in front of the agent;
%\item 
$\atm{dangerous}((X,Y))$ 
%denotes the existence of 
that a dangerous object is at $(X,Y)$;
%\item 
$\atm{xdir}(X)$ and $\atm{ydir}(Y)$ assert the general horizontal and vertical direction, respectively, of the next object on the path in relation to the agent;
%(east, west or on axis).
%\item 
%$\atm{ydir}(Y)$ asserts the general vertical direction of the next object on the path in relation to the agent;
%(north, south or on axis).
%\item 
$\atm{touching}(T)$ 
asserts 
that the next object
$T$ on the path is directly in front of the agent;
and
%denotes the existence of some object that is relevant to the current subtask and located directly in front of the agent;
%and
%This information is needed to decide how the agent should interact with the object (move forward, pick it up, toggle it, etc.).
%\item 
$\atm{ingap}(G)$
%asserts 
that the agent is in a gap (e.g. between rooms).
%, usually between two rooms.
%\end{itemize}

\begin{example}
%Consider the state $s_1$ from Fig.~\ref{fig:ex-rmdp-mg}.
The computed path in $s_1$  (Fig.~\ref{fig:ex-rmdp-mg})
%from the agent to the goal 
is
$(3,1)-(3,5)-(5,3)-(6,6)$. 
Based on the next node on that path, $(3,5)$,
the state is augmented with the facts
$\atm{oriented}(\trm{east})$,
$\atm{fronting}((4,1))$,
$\atm{xdir}(\trm{onAxis})$,
$\atm{ydir}(\trm{south})$,
$\atm{touching}(\trm{none})$, and
$\atm{ingap}(\trm{none})$.
\end{example}

%\paragraph{CARCASS encoding.}
\noindent We 
%define
use 757 abstract states (excluding $\ccs_\infty$): 
%The first 
$\ccs_1$ covers potentially dangerous situations,
%and
%%abstract state $\ccs_1$,
%is 
encoded as 
	\[
		P_{\ccs_1} = \left\{ \begin{array}{lll}
			\atm{sIdx}(\trm{danger}, 1).&&  \\
		  	\atm{sCov}(\trm{danger}, ()) 
				& \leftarrow & \atm{fronting}(T),
                               \atm{dangerous}(T).
			\end{array}\right\},
	\]
while the states $\ccs_2$--$\ccs_{757}$, representing all possible combinations of ground instances for
\atm{oriented}, \atm{xdir}, \atm{ydir}, \atm{touching}, and \atm{ingap},
are compactly encoded as
% a single compact encoding
	\[
		P_{\ccs_{2-757}} = \left\{ \begin{array}{l}
			\atm{label}((\trm{oriented}(O), \trm{xdir}(X), \trm{ydir}(Y), \trm{touching}(T), \trm{ingap}(G))) \\
			\quad \leftarrow \atm{oriented}(O), \atm{xdir}(X), \atm{ydir}(Y), \atm{touching}(T), \atm{ingap}(G). \\
			\atm{sIdx}(S, 2) \leftarrow \atm{label}(S).  \\
		  	\atm{sCov}(S, ())  \leftarrow \atm{label}(S).  \\
			\end{array}\right\}.
	\]
%Intuitively, $\ccs_1$ covers states where the agent is in a potentially dangerous situation (e.g. about to jump in a pool of lava).
%and represent all possible combinations of arguments for $\atm{label}(T)$. 
%The tuple $T$ serves as the label of its respective state.
The background knowledge is designed such that every answer set can have at most one instance of $\atm{label}$, which means that the order restriction can be relaxed in favour of a more compact encoding, 
such that all states $\ccs_2$--$\ccs_{757}$ have the same index in the decision list.
%\paragraph{Abstract Actions.}

While actions can simply be stated as facts, e.g.,
$\atm{aCov}(\trm{left}, \trm{left}).$,
%there is also an opportunity to
we can prune actions that are admissible in the concrete representation but have no effect (i.e., do not lead to state changes). This can be done by omitting such actions (i.e., making them inadmissible) in the abstract representation.
%with the background knowledge 
%it is also possible to prune the set of admissible actions
%by removing ones that do not change the state;
For example,
$
\atm{aCov}(\trm{pickup}, \trm{pickup}) \gets \atm{fronting}(T), \atm{obj}(\trm{key}(\_), T)
$
ensures that the $\atm{pickup}$ action is admissible in the abstract representation only if there is a key to be picked up.
\section{Empirical Evaluation}\label{sec:eval}
%%%%%%%%%%%%%%%%%%%%%%%%%%%%%%%%%%%%%%%%%%%%%%%%%%%%%%%%%%%%%%

%We study the effects of the presented abstractions on abstract {\qlearning}.
We next investigate whether abstractions make large-scale RMDPs tractable 
by reducing
%both the 
space and sample 
%requirements 
needs of the learning process.
%Our primary assumption is that abstractions make large-scale RMDPs tractable 
%by reducing both the space and sample requirements of the learning process.
Notably,
there is no guarantee of convergence of abstract {\qlearning}
%is however not guaranteed
in general 
%and
%if convergence occurs nevertheless, 
%there is also no guarantee
nor that a learned abstract policy is optimal.
Our experiments are thus designed to shed light on {\bf(E1)}
the \emph{stability of the learning process},
{\bf (E2)} the \emph{quality of the learned policy}, and
{\bf(E3)} the \emph{differences in sample efficiency} between concrete and abstract {\qlearning}.

%To understand the effects of our abstraction,
%we therefore study
	%	This leads to our research questions,
	%	formulated in comparison with the concrete RMDP representations (as control condition).
	%		\begin{description}
	%			\item[Q1] Do the proposed abstraction cause differences in sample efficiency?
	%			\item[Q2] Do the proposed abstraction cause stability issues during the learning process?
	%			\item[Q3] Do the proposed abstraction cause differences in quality of the learned policy?
	%		\end{description}

%\paragraph{Experiment setup.}
Both concrete and abstract {\qlearning}
were tested in a $20$-blocks world 
with the goal to stack all blocks in order,
and in three MiniGrid environments.
%\emph{DoorKey $8 \times 8$},
%\emph{Foor Rooms}
%and \emph{MultiRoom N2 S4}.
An $\epsilon$-greedy behaviour policy was used for exploration;
%with different configurations of $\epsilon$, $\alpha$, the initial $q$-values,
for full details, including platform, software, and parameter settings see the supplementary material. %\ref{sec:params}.

%\subsubsection{Metrics and Operationalisation.}
%For studying stability,
As for {\bf(E1)}, we observe the return of episodes across the learning process. Concerning {\bf(E2)}, we measure 
for every realised episode $h_i$ the solution quality by
the \emph{return} $\rv G_0 (h_i)$ at time~0.
We call an episode
\emph{successful} if it achieved a return 
%$\rv G_ 0 ( h_i) > \beta$
above a baseline,
%$\beta$, with 
	viz.\ $\rv G_ 0 ( h_i) > 0$ for MiniGrid and
        $\rv G_ 0 ( h_i) \geq 62$ for Blocks World
        (i.e., the worst-case return of a naive unstack-stack strategy).
	%$\beta = 100 - 2(n-1)$ for an $n$-blocks world environment.
		%This is the worst-case return of a naive unstack-stack policy,
		%requiring $19$ moves to place all blocks on the table and another $19$ to stack
		%them in order.
        %For example, $\beta = 62$ in the $20$-blocks world.
%
For %studying 
measuring {\bf(E3)} sample efficiency,
	we 
    %observe 
    use the 
        \emph{mean return}
        $\frac{1}{n} \cdot \sum_{i=1}^n \rv G _0 (h_i)$
and count the number of successful episodes
over $n$ realised episodes.
    % $\rv H_1, \rv H_2, \ldots, \rv H_n$.
    %\emph{normalised cumulative return} (\textit{NCR})
	%as defined by $\textit{NCR}(h_i) \defined \frac{1}{i} \cdot \sum_{j=1}^i \rv G _0 (h_j)$

\begin{figure}[t] \small
		\includegraphics[width=\textwidth]{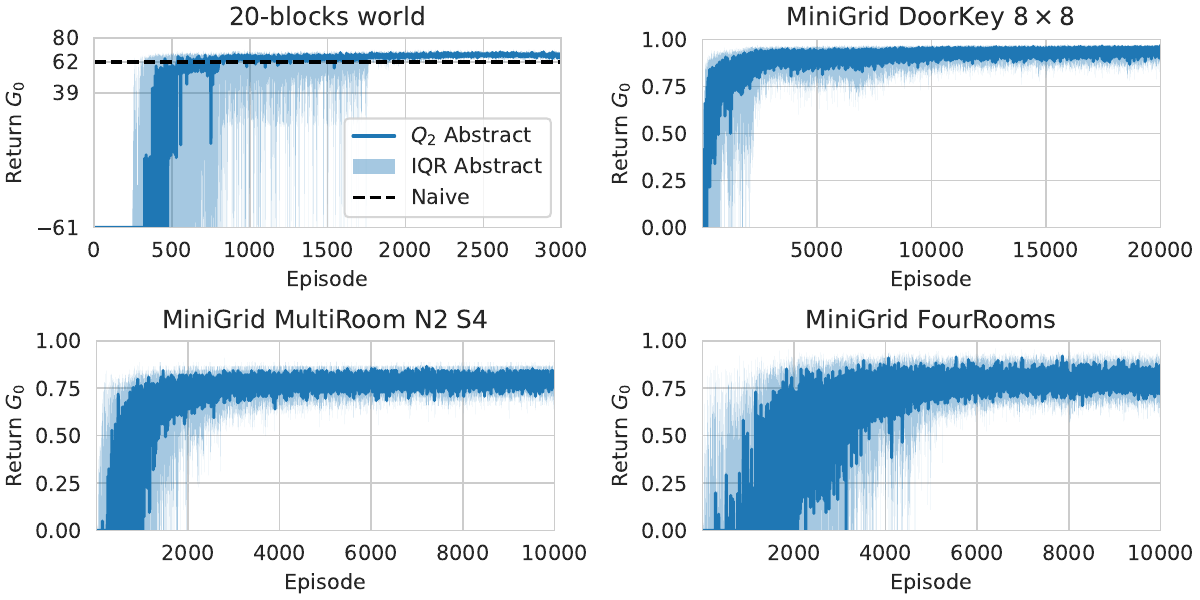}
		\caption{The learning curves of abstract {\qlearning} in four task environments.}
			%	The dark blue line shows the median $Q_2$ and
			%	the light blue area covers the \textit{IQR}.
			%	The baseline for the blocks world environment
			%	is indicated via a dashed line.}
		\label{fig:plt:lc}
	\end{figure}

We repeated experiments 20 times.
The collected data is summarised using
the first quartile $Q_1$, the median $Q_2$, the third quartile $Q_3$, and the interquartile range \textit{IQR}.
%
%\paragraph{Results.}
Fig.~\ref{fig:plt:lc} shows the learning curves of abstract {\qlearning} for the tested environments.
	For the $20$-blocks world,
	the first positive median return is observed
	at episode 376 and after episode 2000 for merely 10 instances $Q_1$ dropped below the baseline of 62, and for none below 60. 
%	For the Door Key, Multi Room, and Four Rooms environments,
For Door Key, Multi Room, and Four Rooms,
	the first positive median returns are observed after episodes 68, 133, and 287, respectively,
	all recorded median returns are positive after episodes 195, 1012, and 3137, respectively,
	and
	all recorded $Q_1$ returns are positive after episodes 2048, 1962, and 4649, respectively.
		%	20 blocksworld
		%		The first positive median return is observed at episode 376,
		%		with median returns being consistently above 62 after episode 1543. After episode 2000,
		%		there are only 10 instances for which Q1 drops below the baseline of 62, and none below
		%		60. 
		%	doorkey
		%		For the variant of
		%		abstract Q-learning with restricted action sets, the first positive median return is recorded
		%		after episode 68. The first positive Q1 return is recorded after episode 122. After episode
		%		195, all recorded median returns are positive. After episode 2048, all recorded Q1 returns
		%		are positive.
		%	
		%	multiroom
		%		Starting out at a return of zero, the first positive median return is
		%		recorded after episode 133. The first positive Q1 return is recorded after episode 639.
		%		After episode 1012, all recorded median returns are positive. After episode 1962, all
		%		recorded Q1 returns are positive. The results for concrete Q-learning, are not depicted
		%		in the plot, as Q1 and Q2 are zero for all episodes. There are exactly two instances for
		%		which Q3 is positive: episodes 5513 and 9803, both with a Q3 return of 0.025.
		%	
		%	fourrooms
		%		Starting out at a return of zero, the first positive median return is
		%		recorded after episode 287. The first positive Q1 return is recorded after episode 1403.
		%		After episode 3137, all recorded median returns are positive. After episode 4649, all
		%		recorded Q1 returns are positive	

Table~\ref{tab:results} 
shows 
(from left to right)
the mean return over all realised episodes and
the percentage of successful episodes over all realised, resp., the last 500 episodes,
%and the percentage of successful episodes over the last 500 episodes
for abstract and concrete {\qlearning} in the four tested environments.
Clear differences can be observed in all environments
when comparing abstract and concrete {\qlearning}.

\begin{table}[t] \footnotesize
\caption{Comparison of abstract and concrete {\qlearning}  (QL) in four task environments.
}\label{tab:results}
\centering
%\begin{tabular}{|c|c||ccc|ccc|ccc|}
\begin{tabular}{ll@{\hspace{.2cm}}rrr@{\hspace{.4cm}}rrr@{\hspace{.4cm}}rrr}
\toprule
%\cline{1-13}
\multirow[l]{2}{*}{\bf Environment} & \multirow[l]{2}{*}{\bf Repr.}   & \multicolumn{3}{c}{\bf Mean $\mathbf{\rv G_0}$  (all eps.)} & %\multicolumn{3}{c|}{\# Successful} & \multicolumn{3}{c|}{\# Successful} \\ 
\multicolumn{3}{c}{\bf\# Succ.~(all eps.)} & \multicolumn{3}{c}{\bf \# Succ.~(last 500)} \\ 
%& & \multicolumn{3}{c|}{(all episodes)} & \multicolumn{3}{c|}{(all episodes)} & \multicolumn{3}{c|}{(last 500)} \\
%& & \multicolumn{3}{l}{\bf (all episodes)} & \multicolumn{3}{l}{\bf (all episodes)} & \multicolumn{3}{l}{\bf (last 500)} \\ 
%\hline
%\cline{3-11}
 &  & $\mathbf{Q_1}$ & $\mathbf{Q_2}$ & $\mathbf{Q_3}$ & $\mathbf{Q_1}$ & $\mathbf{Q_2}$ & $\mathbf{Q_3}$ & $\mathbf{Q_1}$ & $\mathbf{Q_2}$ & $\mathbf{Q_3}$ \\ 
\midrule

20-blocks world & Abstract QL & 18.64 & 43.86 & 54.07 & 0.40 & 0.76 & 0.85 & 0.93 & 0.97 & 0.98 \\
%& Concrete  &-&-&-&-&-&-&-&-&- \\ 
& Concrete QL & -61.00 & -61.00 & -61.00 & 0.00 & 0.00 & 0.00 & 0.00 & 0.00 & 0.00 \\
\midrule
\multirow[t]{2}{*}{Door Key} & Abstract QL & 0.82 & 0.86 & 0.90 & 0.93 & 0.97 & 0.99 & 1.00 & 1.00 & 1.00 \\
 & Concrete QL & 0.01 & 0.01 & 0.01 & 0.02 & 0.02 & 0.03 & 0.02 & 0.02 & 0.04 \\
\midrule
\multirow[t]{2}{*}{Four Rooms} & Abstract QL & 0.59 & 0.61 & 0.64 & 0.78 & 0.81 & 0.84 & 1.00 & 1.00 & 1.00 \\
 & Concrete QL & 0.02 & 0.02 & 0.02 & 0.03 & 0.03 & 0.04 & 0.03 & 0.04 & 0.04 \\
\midrule
\multirow[t]{2}{*}{Multi Room} & Abstract QL & 0.68 & 0.69 & 0.70 & 0.91 & 0.91 & 0.92 & 1.00 & 1.00 & 1.00 \\
 & Concrete QL & 0.01 & 0.01 & 0.01 & 0.02 & 0.02 & 0.02 & 0.02 & 0.03 & 0.03 \\
\bottomrule
\end{tabular}
\end{table}

%\paragraph{Discussion.}

{\bf(E1)} Regarding \emph{stability},
after initial learning,
the returns for abstract $Q$-learning stay consistently high
in all environments, 
with no visible fluctuations in 
the median or the interquartile range. 
Therefore, if stability issues are present, %we conclude that 
the
majority of episodes realised by abstract $Q$-learning are unaffected. 
Still, stability issues may be present in less than 25\% of experiment repetitions.

{\bf(E2)} Regarding \emph{quality},
the high success rates of episodes over the last 500 episodes
suggest that episodes of acceptable quality can be learned
%by abstract {\qlearning}
%for all environments and 
with high consistency.
We cannot make statements about the optimality of the learned policies.
However, for the $20$-blocks world, there are still episodes
where the policy does not match the worst-case return of a naive unstack-stack policy. This may be 
caused by the exploration policy, by other factors (e.g., choice of $\alpha$, more training needed), or by the abstraction itself.

{\bf(E3)} For \emph{sample efficiency},
abstract {\qlearning}
improves on sample efficiency over concrete {\qlearning} 
bases on 
the differences
in the mean returns
and the percentages of successful episodes over the entire learning processes.
%that 
%for the investigated environments. 
%and parameter configurations.

%That is, 
In conclusion, abstract $Q$-learning reliably produced policies of acceptable quality and, moreover, 
\emph{in a smaller number of episodes} 
as compared to concrete $Q$-learning.
%, for all tested tasks and environments.

%	The increase in sample efficiency can be explained
%	in part by the reduction in size of the state-action space due to the abstract representation.
%	\tk
%	As additional factors
%	with a possible effect on sample efficiency, 
%	an exploration bias in the blocks world abstraction and the restricted admissibility of
%	actions in the Minigrid abstraction.

%%%%%%%%%%%%%%%%%%%%%%%%%%%%%%%%%%%%%%%%%%%%%%%%%%%%%%%%%%%%%%
\subsection{Comparison with Prolog-based Abstractions}\label{sec:asp_prolog}
%%%%%%%%%%%%%%%%%%%%%%%%%%%%%%%%%%%%%%%%%%%%%%%%%%%%%%%%%%%%%%
%In addition to the ASP encodings of the presented abstractions, 
%corresponding 
We also developed
Prolog encodings (to be found in the supplementary material) % \ref{sec:enc:bw:prolog}, \ref{sec:enc:mg:prolog})
to contrast the modelling styles encouraged by the two
formalisms for the considered relational decision-making problems.
%The comparison is however not intended as a general assessment of ASP versus Prolog.

The Prolog encoding of the Blocks World CARCASS was relatively simple to obtain, as it is almost identical to the ASP encoding. %, which does not use much ASP-specific Syntax (it is mostly just a normal logic program). 
%Still, during modelling we ran into endless loops already for this very simple encoding, which was resolved by changing the ordering of literals in some rule bodies.
The differences in the modelling approaches become more apparent in the Minigrid abstraction,
which can be viewed as a shortest-path problem under relational constraints.
In the ASP encoding, admissible paths and optimality criteria are specified
declaratively using choice rules, constraints, and optimisation statements,
resulting in a compact encoding that closely follows the problem description.
In the Prolog encoding, solving the same abstraction requires an explicit
implementation of the underlying search process (in our case, a breadth-first
search), leading to a more algorithmic representation in which aspects of
control and problem specification are more explicitly represented.
% Keeping the structure of the encoding as similar as possible to the ASP version, we found that the Prolog encoding suffered from significant performance problems (over a million of proof steps), which could be resolved by adding tabling (below twenty-thousand steps). %Further optimisation is likely possible but we assume that this would hurt the readability of the encoding.
Further details are provided in the supplementary material.% \ref{sec:comp}.

%%%%%%%%%%%%%%%%%%%%%%%%%%%%%%%%%%%%%%%%%%%%%%%%%%%%%%%%%%%%%%
\section{Related Work}\label{sec:rel}
%%%%%%%%%%%%%%%%%%%%%%%%%%%%%%%%%%%%%%%%%%%%%%%%%%%%%%%%%%%%%%
%FoxCS: Memory management for Prolog with tabling

Research on relational reinforcement learning has long explored ways of representing state-action abstractions in first-order logic. 
Prominent examples include \emph{logical Markov decision programs} (LOMDPs)~\citep{kersting_raedt_2004}, which describe abstract states as logical rules and transitions as probabilistic effects of actions, and \emph{relational $Q$-learning}~\citep{morales_2003}, which groups states and actions into logical templates that define when an abstract action can be applied. 
Other extensions include \emph{probabilistic relational models} (PRMs)~\citep{guestrin2003planning}, which decompose value functions across object classes,
%and 
work on \emph{stochastic policies for relational POMDPs}~\citep{itoh2004towards}, where policies are expressed as decision lists with probabilities,
and a learning classifier system in first-order logic (FOXCS)~\citep{mellor_2008}.
All of these approaches share with \emph{CARCASS}~\citep{otterlo_2008_phd} the goal of reducing the state explosion problem through logical abstraction, but they differ in their underlying mechanisms: LOMDPs match concrete states against general logical rules, $\mathit{RQ}$-learning defines partitions of the state-action space through conjunctions of relations, and PRMs assume additive decompositions over objects. 
CARCASS instead represents abstractions as logical rules with background knowledge, implemented in Prolog using SLDNF resolution. 
Like CARCASS, FOXCS represents policies via sets of rules, implemented in Prolog, that can generalise over states and actions. Unlike CARCASS, the head of each rule consists of exactly one abstract action (not multiple) and the set of rules is not interpreted as a decision list, but individual rules advocate for their covered actions in a weighted voting process.
Our work continues the CARCASS line of research but replaces Prolog with a fully declarative ASP encoding, which allows richer forms of reasoning, including nonmonotonic inference and optimisation, within the abstraction process.

Several works also combine \emph{ASP and reinforcement learning} %, though in very different ways.
but quite differently. 
For instance, relational meta-policies have been encoded in ASP for hierarchical option selection~\citep{mitchener_2022}, and other
%approaches 
works use ASP to restrict exploration by pruning infeasible actions, e.g.~\citep{santos_ferreira_2017}. 
%In robotics, ASP has been employed for reasoning about incomplete knowledge, planning, or diagnosis~\citep{yang_2014,zhang_2015,sridharan-meadows-2018,mota-sridharan-leonardis-2021}. 
ASP %has also been 
was also used to encode entire relational MDPs, including states and value functions~\citep{saad_2011}. 
However, none of these works proposed a CARCASS-style general ASP-based representation of \emph{full state-action abstractions},
%in the style of CARCASS, 
which is 
%the focus of 
our core contribution.

\section{Conclusion}\label{sec:concl}
%%%%%%%%%%%%%%%%%%%%%%%%%%%%%%%

We investigated ASP for modelling state-action pair abstractions in relational MDPs, building on the CARCASS framework. We presented a general encoding method and evaluated it on Blocks World and MiniGrid tasks. In both domains, abstract representations enabled {\qlearning} to learn high-quality policies consistently with fewer samples than concrete representations, showing its potential as a knowledge-driven framework for relational RL abstractions that supports elaboration-tolerant policy learning.
ASP-based abstractions support generalisation, action restrictions, and problem decomposition by allowing declarative encoding of domain knowledge. They also support nonmonotonic reasoning, optimisation, and reasoning with incomplete knowledge, providing a flexible alternative to Prolog-based CARCASS abstractions. %Beyond the case studies, this suggests that ASP can complement statistical relational approaches by injecting symbolic knowledge into RL.
%,  and potentially improving sample efficiency and generalisation.

Future work includes automatic generation and refinement of ASP abstractions~\citep{saribatur-schueller-eiter-2019}, leveraging nondeterminism and optimisation to guide exploration, and testing transfer of abstract policies to larger or previously unseen tasks. 
%Improving computational efficiency via multi-shot solving and assumptions~\citep{gebser-kaminski-kaufmann-schaub-2018-multishot} is also a key avenue for scaling to more complex environments. 
Additional directions include exploring elaboration-tolerant design patterns, integrating ASP abstractions with function approximation methods, and evaluating the benefits of symbolic priors for RL in richer relational settings.

    %The original RMDP definition uses the \emph{Herbrand atom base} (not the \emph{Herbrand literal base}) and operates under
	%the closed world assumption (where atoms not in $s \in S$ are assumed to be \emph{false}).
	%We adapt this version to stay compatible with the CARCASS Framework, which is based on normal logic programs.
	%Allowing strong negation in $S$ does however make sense and could be very useful in a partially observable context. 
	%It is not immediate whether allowing strong negation in $A$ (i.e. the inclusion of strongly negated actions) makes sense.

	%	\tk We assume a standard Herbrand interpretation with the closed world assumption.
	%	But it may be fruitful to 
	%	interpret states as \emph{belief states},
	%	where an atom $\atm{p}(\bar t)$ is \emph{true} in $s$ if $\atm{p}(\bar t) \in s$,
	%	\emph{false} in $s$ if $\neg \atm{p}(\bar t) \in s$, and \emph{unknown} in $s$ otherwise.
	%	This and future definitions was originally given in a normal logic programming setting, where strong negation is not part of the language. Therefore we need to exclude it.
	%	But we see no harm in including strongly negated predicate atoms in abstract states when encoded in ASP.
	%	Also, built-in atoms (e.g. equality) can be included to some extent,
	%	as was done in examples by \citep[p.~253]{otterlo_2008_phd} and in  Section~\ref{sec-carcass-example}
	%	but one needs to be careful when translating them from Prolog to ASP.

\section*{Acknowledgements}

This work has been supported by funding from the Knowledge Foundation (KK-stiftelsen) within the synergy project \emph{Efficient and Trustworthy Industrial AI} (ETIAI).

\section*{Competing interests}
The authors declare none.

%%%%%%%%%%%%%%%%%%%%%%%%%%%%%%%%%%%
%%%%%%%%%%%%%%%%%%%%%%%%%%%%%%%%%%%
{ \small
\bibliographystyle{abbrvnat}
%\bibliography{refs_short}
%\input{references}
\bibliography{refs}
}

\end{document}

%% file: prop-correctness.tex
Given an RMDP $M = \rmdp$, a CARCASS $\abs C$,
        and labels $\lambda$,
        let 
		$s \in S$,
		$P= P_s \cup P_{A_s} \cup P_\textit{Cov}$,
		%Let $P = P_s \cup P_{A_s} \cup P_{\abs\Psi}$, 
		$I \in \as (P)$, and
		$\ccs_i = l_1, \ldots, l_k \in \abs S$ with variables $V_1, \ldots, V_m$.
		%and $\lambda \fdef \abs S \cup \abs \Psi \mapsto \setofstrings$.
		Furthermore, let $\theta $ be a substitution such that $\ccs_i \theta$ is a ground instance of $\ccs_i$.
		Then, $P_s \sldnf \ccs_i \theta$ iff $\atm{sCov}(\lambda(\ccs_i), (V_1 \theta, \ldots, V_m \theta)) \in I$. 